\documentclass[pdflatex,sn-mathphys-num,iicol]{sn-jnl}

\usepackage{graphicx}%
\usepackage{multirow}%
\usepackage{amsmath,amssymb,amsfonts}%
\usepackage{amsthm}%
\usepackage{mathrsfs}%
\usepackage[title]{appendix}%
\usepackage{xcolor}%
\usepackage{textcomp}%
\usepackage{manyfoot}%
\usepackage{booktabs}%
\usepackage{algorithm}%
\usepackage{algorithmicx}%
\usepackage{algpseudocode}%
\usepackage{listings}%

\usepackage[braket, qm]{qcircuit}
\usepackage{tikz}
\usetikzlibrary{shapes.geometric, arrows.meta, positioning}
\usetikzlibrary{shapes.geometric, positioning, fit}
\usepackage{subcaption}
\usepackage{graphicx}
\usepackage{booktabs}
\usepackage{graphicx}
\usepackage{subcaption}
\usepackage{color}
\usepackage{colortbl}
\usepackage{comment}

\theoremstyle{thmstyleone}%

\theoremstyle{thmstyletwo}%

\theoremstyle{thmstylethree}%

\begin{document}

\title[Article Title]{Applying Quantum Autoencoders for Time Series \\ Anomaly Detection}

\author*[*]{\fnm{Robin}  \sur{Frehner}}\email{frehner.robin@gmail.com}
\author*{\fnm{Kurt} \sur{Stockinger}}\email{kurt.stockinger@zhaw.ch}

\affil{
\orgname{Zurich University of Applied Sciences}, 
\country{Switzerland}}



\abstract{Anomaly detection is an important problem with applications in various domains such as fraud detection, pattern recognition or medical diagnosis. Several algorithms have been introduced using classical computing approaches. However, using quantum computing for solving anomaly detection problems in time series data is a widely unexplored research field.

This paper explores the application of quantum autoencoders to time series anomaly detection. We investigate two primary techniques for classifying anomalies: (1) Analyzing the reconstruction error generated by the quantum autoencoder and (2) latent representation analysis. Our simulated experimental results, conducted across various ansaetze, demonstrate that quantum autoencoders consistently outperform classical deep learning-based autoencoders across multiple datasets. Specifically, quantum autoencoders achieve superior anomaly detection performance while utilizing 60-230 times fewer parameters and requiring five times fewer training iterations. In addition, we implement our quantum encoder on real quantum hardware. Our experimental results demonstrate that quantum autoencoders achieve anomaly detection performance on par with their simulated counterparts. }



\maketitle

\section{Introduction}
\label{sect:introduction}

Anomaly detection in time series is a critical task across various domains, including network monitoring, medical diagnosis, and financial fraud detection \cite{Blazquez,AnomalyDetectionIOTSurvey,AnomalyDetectionUsingAE,AnomalyDetectionBasedonConvolutionalRecurrentAutoencoder,frehner2024detecting}.  The challenges in this field are multifaceted, with a significant obstacle being the scarcity of anomalies \cite{Blazquez, 9523565, revisitingTimeSeries}. Traditional techniques for event classification often fall short when faced with the imbalance between anomalous and normal events~\cite{AnomalyDetectionIOTSurvey}. Furthermore, the increasing volume of data necessitates novel approaches. 

Recent advancements in quantum computing \cite{bova2021commercial, bravyi2022future, simoes2023experimental, kim2023evidence, kittelmann2024qardest} and in particular in quantum machine learning \cite{ huang2021power, ngairangbam2022anomaly, simoes2023experimental, jerbi2023quantum}, with its demonstrated expressive power, present new opportunities to address the complexities of time series anomaly detection. In this paper, we investigate the applicability of quantum autoencoders \cite{RomeroQAE, BravoQAE} and demonstrate their effectiveness compared to classical autoencoders ~\cite{AnomalyDetectionUsingAE} using a set of challenging benchmark datasets.

The quantum autoencoder introduced by \cite{RomeroQAE, BravoQAE} shares the same idea with classical deep learning-based autoencoders where the dimension of the input time series is reduced by an \textit{encoder} and then the compressed representation can be decoded by a \textit{decoder} to produce a model output. Similar to deep learning-based autoencoders, anomalies can be identified by analyzing the latent representation obtained after encoding the input with the quantum autoencoder, or by measuring the difference between the model output and the original input. This approach is highly flexible, can integrate various techniques from other anomaly detection methods and is particularly suited for challenging anomaly detection problems where only normal events are available for training, while the actual anomalies are only contained in the test set -- which is an extreme form of scarcity of anomalies. An example of such a challenging dataset is the benchmark from University of California, Riverside (UCR)\cite{TSBenchmarkFlaws}.

This paper presents a comprehensive investigation of the application of quantum autoencoders for time series anomaly detection. The main contributions of this work are as follows:

\begin{enumerate}
    \item We detail the use of quantum autoencoders for time series anomaly detection, establishing the framework for their application.
    \item We employ two methods for anomaly classification: reconstruction error and latent representation analysis.
    \item We demonstrate our approach through simulations, showing quantum autoencoders outperform classical deep learning-based autoencoders on every employed data set with 60-230 times fewer parameters and requiring 5 times fewer training iterations.
    \item We successfully demonstrate the implementation of quantum autoencoders on real quantum hardware, achieving anomaly detection performance comparable to its simulated counterpart, despite the presence of quantum noise.
\end{enumerate}






The paper is structured as follows. In Section \ref{sect:preliminaries}, we provide an overview of the taxonomy of anomalies in time series and describe the quantum autoencoder proposed by \cite{RomeroQAE, BravoQAE}. In Section \ref{sect:methodology}, the overall methodology is presented in detail. The specific experimental setup, including the dataset utilized, evaluation metrics employed, methodology for classification, and parameters employed to conduct and reproduce the experiments, is outlined in Section \ref{sect:experimental-setup}. The experimental results are analyzed in Section \ref{sect:results}. Related work is discussed in Section \ref{sect:relatedWork}. Finally, the paper concludes with a summary of the findings and future considerations in Section \ref{sect:DiscussionAndConclusion}.

\section{Preliminaries}
\label{sect:preliminaries}
This section presents the foundational concepts necessary for this study, organized into two key subsections. The first subsection, Time Series Anomalies, introduces the concept and classification of anomalies in time series data, emphasizing the challenges associated with their detection. The second subsection, details the architecture, operational principles, and application of quantum autoencoders for anomaly detection in time series data.
 \begin{figure*}
\begin{subfigure}{.33\textwidth}
  \centering
  \includegraphics[width=1\linewidth]{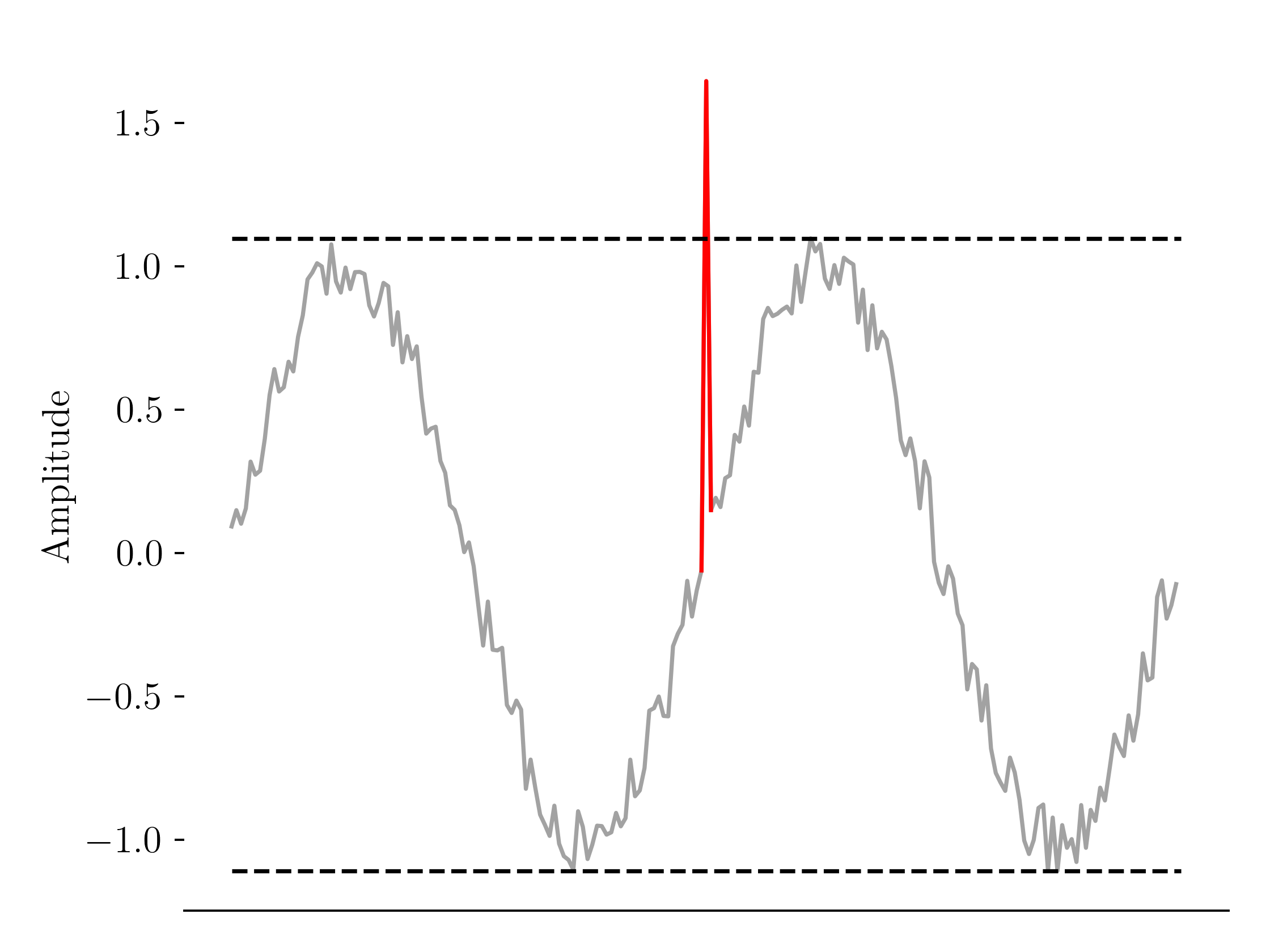}
  \caption{Point Anomaly}
  \label{fig:AnomalyTypesSfig1}
\end{subfigure}%
\begin{subfigure}{.33\textwidth}
  \centering
  \includegraphics[width=1\linewidth]{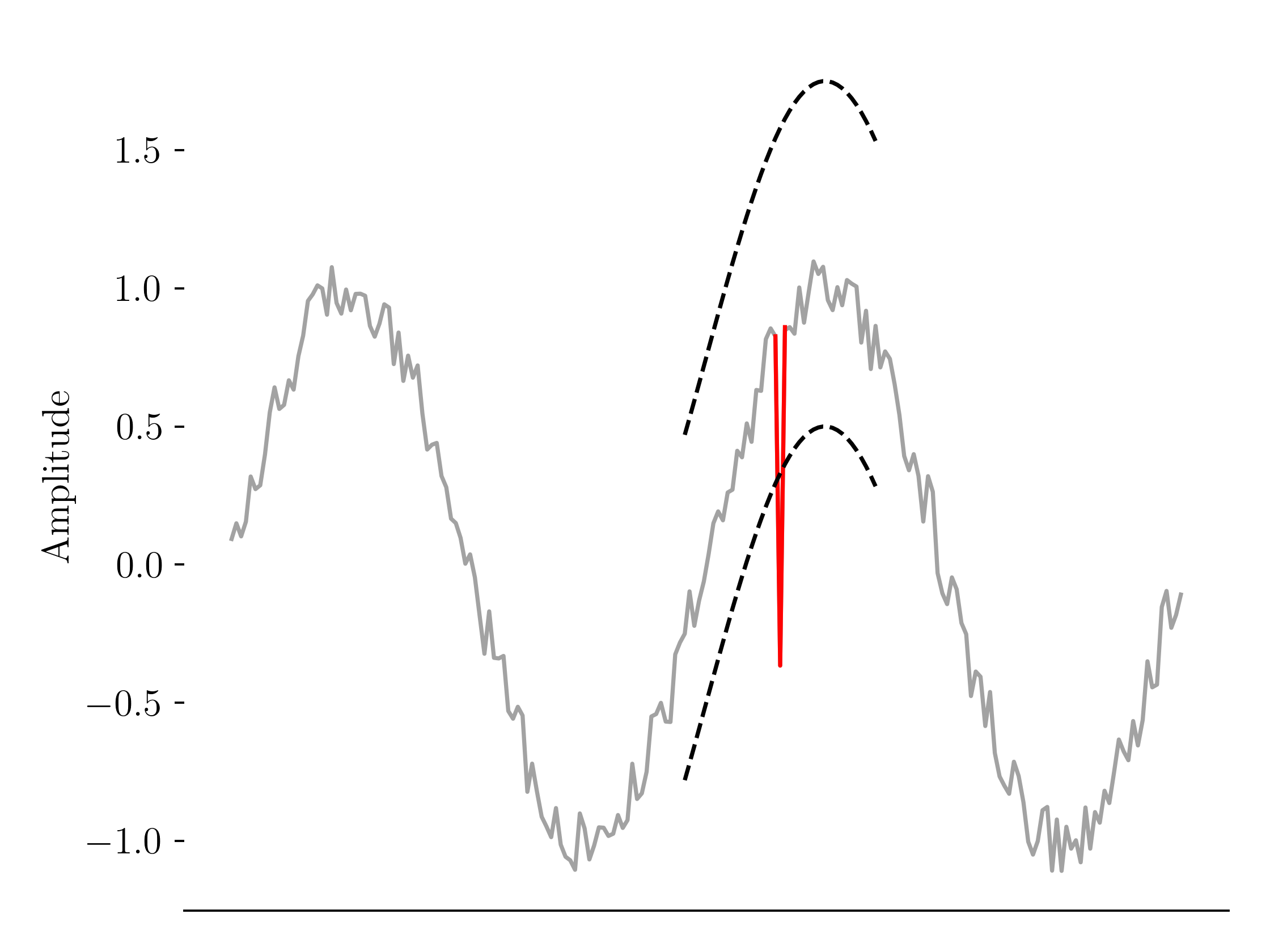}
  \caption{Contextual Anomaly}
  \label{fig:AnomalyTypesSfig2}
\end{subfigure}
\begin{subfigure}{.33\textwidth}
  \centering
  \includegraphics[width=1\linewidth]{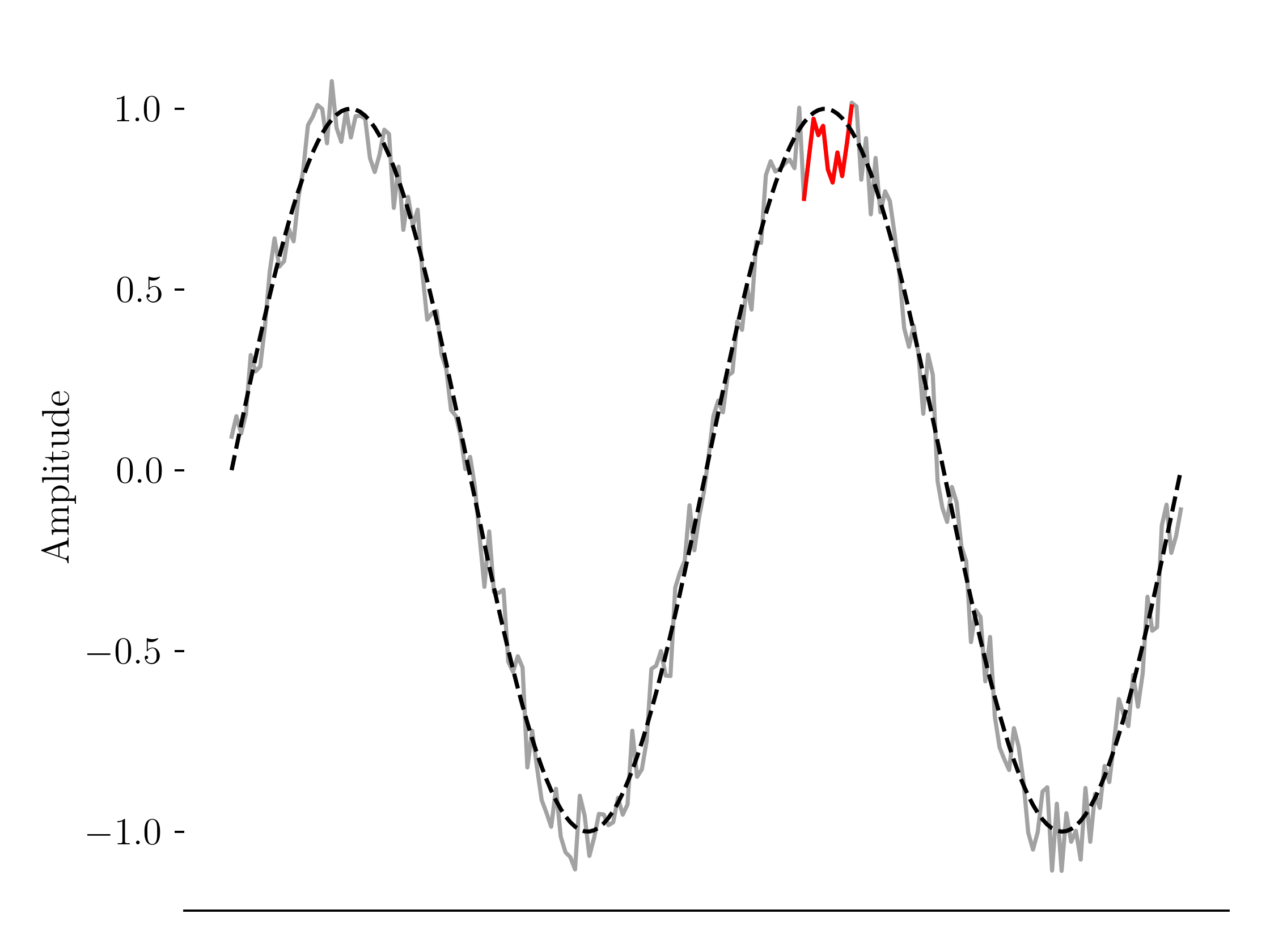}
   \caption{Collective Anomaly}
  \label{fig:AnomalyTypesSfig3}
\end{subfigure}

\caption{Examples of different anomalies in time series data. Dashed lines indicate normality thresholds, red lines highlight anomalies, and grey lines represent the observed time series. Figure \ref{fig:AnomalyTypesSfig1} shows a point anomaly, where a value of 1.5 exceeds the normal range. Figure \ref{fig:AnomalyTypesSfig2} illustrates a contextual anomaly, where the value is typical but anomalous given its context. Figure \ref{fig:AnomalyTypesSfig3} depicts a collective anomaly, where data points are individually within normal noise levels but collectively exhibit a prolonged period of lower values.}
\label{fig:AnomalyTypes}
\end{figure*}

\subsection{Time Series \& Anomalies Foundation}
A time series, represented as \(X \equiv \{x_i\}\) , is a sequence of values recorded at different time points, where $x_i$ denotes the value at time $i$. For simplicity, $x_i$ is described as a scalar value, although it can be a vector. The terms "data point" or "point" refer to $x_i$, while a "window" represents a contiguous subsequence of $X$, denoted as $\{x_i, x_{i+1}, ..., x_{i+w}\}$, where $w$ represents the window width.

The sliding window approach is commonly employed in time series analysis, dividing the time series into fixed-length segments or windows. Each window is analyzed separately, with the width $w$ determining the number of data points in each segment. By sliding the window along the time series with a step size, overlapping segments are created, enabling a more comprehensive understanding of the data. The window width and step size can be adjusted to suit the specific characteristics of the data being studied. Anomaly detection in time series data involves identifying deviations from expected behavior. Three types of anomalies have been identified in the literature \cite{9523565, revisitingTimeSeries}:

\paragraph{Point Anomalies} 
These anomalies occur at individual time points and significantly deviate from the general pattern or trend of the time series data. They can be spikes or glitches compared to neighboring points \cite{revisitingTimeSeries}. A depiction can be found in Figure \ref{fig:AnomalyTypesSfig1}.

\paragraph{Contextual Anomalies}
Contextual anomalies, shown in Figure \ref{fig:AnomalyTypesSfig2}, refer to data points or sequences observed within a short time window that do not deviate from the normal range in a predefined manner, like point anomalies. However, within the given context, they exhibit deviation from the expected pattern or shape, making them challenging to detect \cite{9523565}.

\paragraph{Collective Anomalies} 
Collective anomalies, illustrated in Figure \ref{fig:AnomalyTypesSfig2}, occur over an extended period and represent a set of data points that collectively deviate from normal patterns. Detecting such anomalies is challenging as they may not be immediately apparent and require examination of long-term context \cite{9523565,revisitingTimeSeries}.

\subsection{Quantum Autoencoder}
\label{sect:QuantumAutoencoder}
\begin{figure*}
\centering
\begin{subfigure}{0.45\textwidth}
\resizebox{1.0\textwidth}{!}{%
  \input{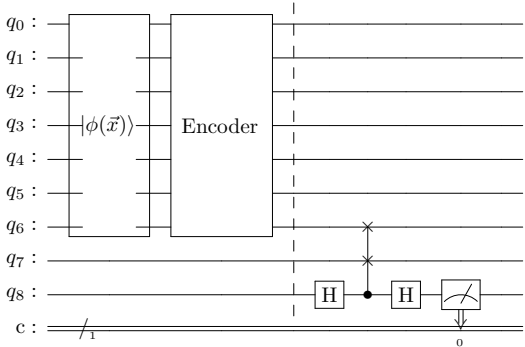}
  }
  
  \caption{Trainable quantum autoencoder architecture}
  \label{fig:QAEArchitectureSfig1}
\end{subfigure}%
\begin{subfigure}{0.55\textwidth}
  \centering
\resizebox{0.925\textwidth}{!}{%
  \input{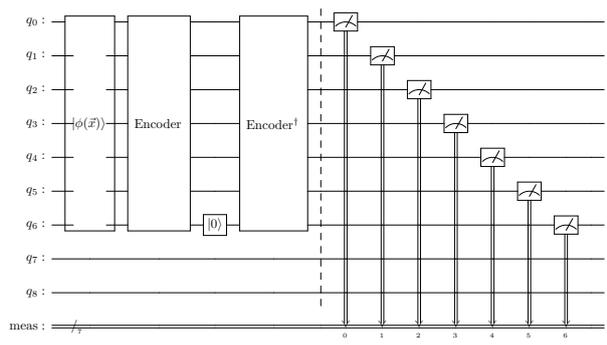}
  }
  \caption{Trained quantum autoencoder architecture}
  \label{fig:QAEArchitectureSfig2}
\end{subfigure}
\caption{The figures depict a 7-qubit quantum autoencoder circuit. Despite requiring 9 qubits for training, it is termed a 7-qubit autoencoder due to the number of qubits in the state preparation and encoder sub-circuit. Figure \ref{fig:QAEArchitectureSfig1} illustrates the circuit used during training, while Figure \ref{fig:QAEArchitectureSfig2} shows the final autoencoder capable of reconstructing input states. Encoder parameters optimized during training the archictecure in Figure \ref{fig:QAEArchitectureSfig1} are transferred to the setup in \ref{fig:QAEArchitectureSfig2}. In both figures, $\phi(\vec{x})$ represents the state preparation procedure, and $Encoder$ denotes a quantum variational sub-circuit with consistent architecture in both setups. Additionally, $|0\rangle$ signifies qubit reset (qubit 6), and $Encoder^{\dag}$ denotes the conjugate transpose of the encoder, serving as the decoding component.}
\label{fig:QAEArchitecture}
\end{figure*}

\cite{RomeroQAE} and \cite{BravoQAE} describe in their work a specific quantum machine learning model designed to perform unsupervised learning tasks, particularly data compression and representation learning, on quantum systems. The described quantum autoencoder builds upon the concept of classical autoencoders and adapts it to quantum computing. In this work, we utilize the implementation proposed by \cite{BravoQAE}, which introduces a two-step architecture employing two dedicated circuits. While the final architecture, shown in Figure \ref{fig:QAEArchitectureSfig2}, is similar to deep learning-based approaches in that it efficiently projects high-dimensional data into a lower-dimensional space, the training process in quantum computing differs. It requires an additional circuit (Figure \ref{fig:QAEArchitectureSfig1}) that is discarded after the training phase is completed.


In this particular case, the quantum autoencoder uses 7 qubits where qubit 0-5 represent the latent space and qubit 6 is called \textit{trash state} as it will be reset - or discarded -  after encoding the quantum state $|\varphi_x\rangle = \varphi(\vec{x})$. The different layers are described in more depth below.

\paragraph{State Preparation} The procedure $\phi(\vec{x})$ transforms input data into the initial quantum state. In this specific instance, the feature space resides in the $2^7 = 128$-dimensional Hilbert space $\mathcal{H}^{128}$. It is crucial to note that state preparation focuses solely on producing a valid quantum state and does not concern itself with reconstructability. When preparing a quantum state for a quantum autoencoder, one must be aware of the fact that, although the prepared quantum state may contain negative values, the output of the circuit comprises measurements that represent a probability distribution, with all its inherent constraints. Consequently, the reconstruction post-measurement will not include any negative values. This aspect is significant in anomaly detection settings, where the identification of anomalies is based on the dissimilarity between the input and the reconstructed, measured output.

\paragraph{Encoder} The encoder block is built upon multiple variational quantum gates and is applied to the initial quantum state to map the data to a lower dimension, resulting in compression. In the illustration, the encoder was trained to map from  a $2^7$-dimensional Hilbert-Space to $2^6$. Ideally, qubit 6 does not hold any information (i.e it is equal to the zero state) after application of the encoder resulting in all of the information condensed into qubits 0 to 5.

\paragraph{Qubit Reset}
In this step, the qubits corresponding to the trash states are reset to the zero state to condense all information into the qubits representing the latent space. This process produces the encoded data, which can be represented with a reduced number of qubits. As demonstrated in Figure \ref{fig:QAEArchitecture}, qubits 0-5 represent the latent space, while qubit 6 corresponds to the trash state and will be reset.

\paragraph{Decoding} The unitary property of the encoding transformation removes the necessity of training a decoder, in contrast to classical deep learning-based autoencoders. The decoder, in this context, becomes a straightforward inversion of the transformation accomplished by applying the compressed data to the conjugate transpose of the encoder. The obtained quantum state in the ideal case is equal to the quantum state after applying the state preparation procedure.

 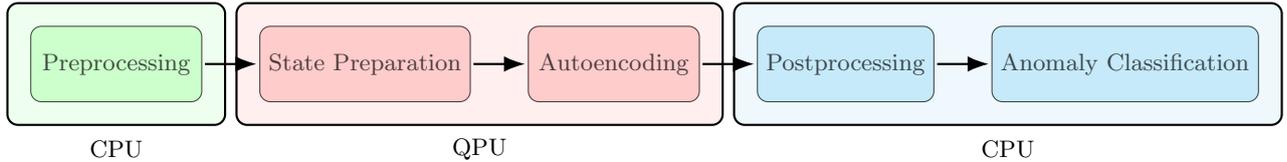
\begin{figure*}
    \centering
    \begin{tikzpicture}[
    node distance=1.5cm and 1.5cm,
    every node/.style={rectangle, rounded corners, minimum width=2.25cm, minimum height=0.6cm, text centered, draw=black, fill=blue!10, font=\small},
    process/.style={rectangle, rounded corners, minimum width=2.25cm, minimum height=1.0cm, text centered, draw=black, fill=blue!10, font=\small},
    box/.style={rectangle, draw=black, thick, inner sep=0.3cm, font=\small},
    arrow/.style={-{Latex[scale=1.2]}, thick},
    startstop/.style={ellipse, minimum width=2.25cm, minimum height=0.6cm, text centered, draw=black, fill=red!20, font=\small}
]

\node (preprocessing) [process, fill=green!20] {Preprocessing};
\node (stateprep) [process, right=0.75cm of preprocessing, fill=red!20] {State Preparation};
\node (autoencoding) [process, right=0.75cm of stateprep, fill=red!20] {Autoencoding};
\node (postprocessing) [process, right=0.75cm of autoencoding, fill=cyan!20] {Postprocessing};
\node (classification) [process, right=0.75cm of postprocessing, fill=cyan!20] {Anomaly Classification};

\node [box, fit=(preprocessing), label=below:CPU, fill=green!20, fill opacity=0.3] {};
\node [box, fit=(stateprep) (autoencoding), label=below:QPU, fill=red!20, fill opacity=0.3] {};
\node [box, fit=(postprocessing) (classification), label=below:CPU, fill=cyan!20, fill opacity=0.3] {};

\draw [arrow, shorten >=1pt, shorten <=1pt] (preprocessing) -- (stateprep);
\draw [arrow, shorten >=1pt, shorten <=1pt] (stateprep) -- (autoencoding);
\draw [arrow, shorten >=1pt, shorten <=1pt] (autoencoding) -- (postprocessing);
\draw [arrow, shorten >=1pt, shorten <=1pt] (postprocessing) -- (classification);

\end{tikzpicture}
    \caption{The high level methodology employed for anomaly detection in this study encompasses a systematic approach, beginning with data preprocessing. Subsequently, the quantum state preparation procedure is performed, followed by applying the chosen autoencoding ansatz. Upon completion of these computational phases, the resultant data is subjected to postprocessing procedures and final anomaly classification. The notations CPU and QPU delineate between actions executed on classical computing architectures (CPU) and those performed on quantum hardware platforms (QPU).}
    \label{fig:MethodologyOverview}
\end{figure*}

\paragraph{Training}
\label{sect:QAEArchitectureTraining}

Resetting of the trash qubit(s) during the encoding process leads to the loss of all information stored in the corresponding qubits and influences the decoding process. Therefore, to mitigate such information loss, an ideal scenario entails the encoder to efficiently map all relevant information to the qubits associated with the latent space (i.e. qubits 0-5 as depicted in Figure \ref{fig:QAEArchitecture}) before the resetting operation occurs. This approach aims to minimize the impact of information loss during the encoding procedure.

The mapping is accomplished using the circuit depicted in Figure \ref{fig:QAEArchitectureSfig1}, which incorporates the state preparation function $\phi(\vec{x})$ and the encoder circuit of the desired autoencoder. This circuit is extended by introducing a reference qubit (qubit 7) initialized as $|0\rangle$ and an ancillary qubit (qubit 8) for executing the Swap-Test \cite{foulds2021controlled}, which determines whether a pair of states are inequivalent. Upon measuring qubit 8, if qubits 6 and 7 are equal, the probability of obtaining a measurement outcome of 0 for qubit 8 is $1$, whereas if qubits 6 and 7 are orthogonal, the probability of observing a 0 for qubit 8 is $\frac{1}{2}$. The primary objective is to train the encoder to effectively map all information onto the qubits representing the latent space, specifically qubits 0-5 in the circuit shown in Figure \ref{fig:QAEArchitectureSfig2}. Achieving this requires maximizing the number of zeros observed in qubit 8, indicating that qubit 6 closely aligns with the initialized $|0\rangle$ state of qubit 7. 

\section{Methodology of Quantum Autoencoders for Anomaly Detection}
\label{sect:methodology}

This section outlines the methodology for applying quantum autoencoders to time series anomaly detection depicted in Figure \ref{fig:MethodologyOverview}. The approach includes preprocessing, state preparation, autoencoding, postprocessing, and anomaly classification. State preparation and autoencoding are done according to \cite{BravoQAE}.

\subsection{Preprocessing}
\label{sect:MethdologyPreprocessing}
Upon considering a time series, we generate sliding windows of a fixed size, specifically $2^K$, where $K$ corresponds to the number of qubits incorporated in the quantum autoencoder, as outlined in Section \ref{sect:QuantumAutoencoder}. Notably, at this stage, we refrain from normalizing the time series data since this operation is subsequently performed during the quantum state preparation process, as elaborated upon in Section \ref{sect:StatePreparation}. Nevertheless, it is important to note that optional normalization of the time series can be carried out if deemed appropriate as in the case of using a classical deep learning-based autoencoder.

\subsection{State Preparation}
\label{sect:StatePreparation}
For state preparation, we adopt \textit{amplitude encoding}, wherein each window $\vec{w_i}$ is transformed into a quantum state $|\varphi_i\rangle$ by leveraging probability amplitudes.We selected amplitude encoding due to its efficiency in encoding a large number of input features with a minimal qubit requirement. Specifically, this method requires only $log(N)$ qubits to encode $N$ datapoints, which is significantly more efficient compared to alternative techniques such as angle encoding. This encoding method establishes a direct correspondence between the input and the probability amplitudes of the $2^K$ qubit state. 

Specifically, given a window $\vec{w_i} = [w_{i0}, w_{i1}, \ldots, w_{i2^K-1}]$ of size $2^K$ spanning from index $i$ to $i + 2^{K-1}$ of the time series, the prepared quantum state using amplitude encoding is expressed as $|\varphi_i\rangle = \frac{1}{\langle \vec{w_i}|\vec{w_i}\rangle} \vec{w_i}$ . It is important to highlight that the state preparation step does not involve trainable parameters for autoencoding and is entirely deterministic in nature. 


\subsection{Autoencoding}
\label{sect:MethodologyAutoencoding}
After state preparation, the time windows are autoencoded using either the trainable or trained architecture shown in Figure \ref{fig:QAEArchitectureSfig1} and Figure \ref{fig:QAEArchitectureSfig2}, respectively. Utilizing the trained architecture, which involves the inverse transform after resetting the trash qubits, resembles classical deep learning counterparts, allowing for the reconstruction of the input and subsequent classification based on the reconstructed time windows.
For anomaly detection without input reconstruction, the trainable setup using only the Swap-Test can be employed, as discussed in Section \ref{sect:QAEArchitectureTraining}. Here, the number of measured 1s should be minimal for benign data observed during training, in contrast to unseen anomalies.

\subsection{Postprocessing}
\label{sect:MethodologyPostprocessing}
Upon autoencoding the initial quantum state $|\varphi_i\rangle$ and deriving the output state $|\varphi_i^{\prime}\rangle$, we either compute the mean squared error $\epsilon_i = \frac{1}{2^K}\langle(\varphi_i - \varphi_i^{\prime})|(\varphi_i - \varphi_i^{\prime})\rangle$ between input and reconstructed output in case we use the autoencoder architecture depicted in Figure \ref{fig:QAEArchitectureSfig2}. If we use the Swap-Test measurements (i.e the trainable setup shown in Figure \ref{fig:QAEArchitectureSfig1}) we generate a vector $\vec{\epsilon} = [\epsilon_i ... \epsilon_i]$ of the same length as the number of time windows we processed, where $\epsilon_i$ is the Swap-Test measurement for the particular time window starting at $t_i$.

Moving averaging is employed to eliminate potential outliers and assumes that for time windows containing anomalous data points the respective metric employed consistently differs from time windows comprising benign data. In this paper we adopt a moving average approach with fixed window size on $\vec{\epsilon}$, yielding $|\vec{\epsilon}| - |M_w|$ estimations for each data point, where $|M_w|$ is the size of the moving average window employed. The resultant post-processed time series is denoted as $P$, where $p_i$ corresponds to the post-processed reconstruction error (or Swap-Test measurement) at time step $i$.

\subsection{Classification}
\label{sect:MethodologyClassification}
The final step involves classifying the time series, identifying anomalies based on the post-processed data as outlined in Section \ref{sect:MethodologyPostprocessing}. Commonly, a threshold-based classification approach is used, wherein a time step $i$ is designated as anomalous if its corresponding post-processed reconstruction error $p_i$ surpasses or equals a pre-defined threshold $t$. The classification procedure used for this study is explained in Section \ref{sect:UCRDataset}.

\section{Experimental Setup}
\label{sect:experimental-setup}

The primary objective of this research is to assess the effectiveness of quantum autoencoders in detecting anomalies in time series data, as well as to demonstrate their successful implementation on real quantum hardware. 

The first experiment evaluates the performance of quantum autoencoders using quantum simulators compared to a deep learning-based baseline as detailed in Section \ref{sect:SimulatorSetup}. The second experiment evaluates quantum autoencoders on real quantum hardware as described in Section \ref{sect:RealQuantumHardwareExperiment}. 


\subsection{Data Set}
\label{sect:UCRDataset}

For our experiments we use the University of California at Riverside (UCR) Time Series Anomaly Archive \cite{TSBenchmarkFlaws}, a benchmark dataset specifically designed for time series anomaly detection. This archive provides a diverse array of datasets from various domains, including medicine, sports, entomology, industry, space science, robotics, and more. Each dataset contains only a single anomaly in the test set and no anomalies in the training data.



\begin{figure*}
\begin{subfigure}{.33\textwidth}
  \centering
  \includegraphics[width=1\linewidth]{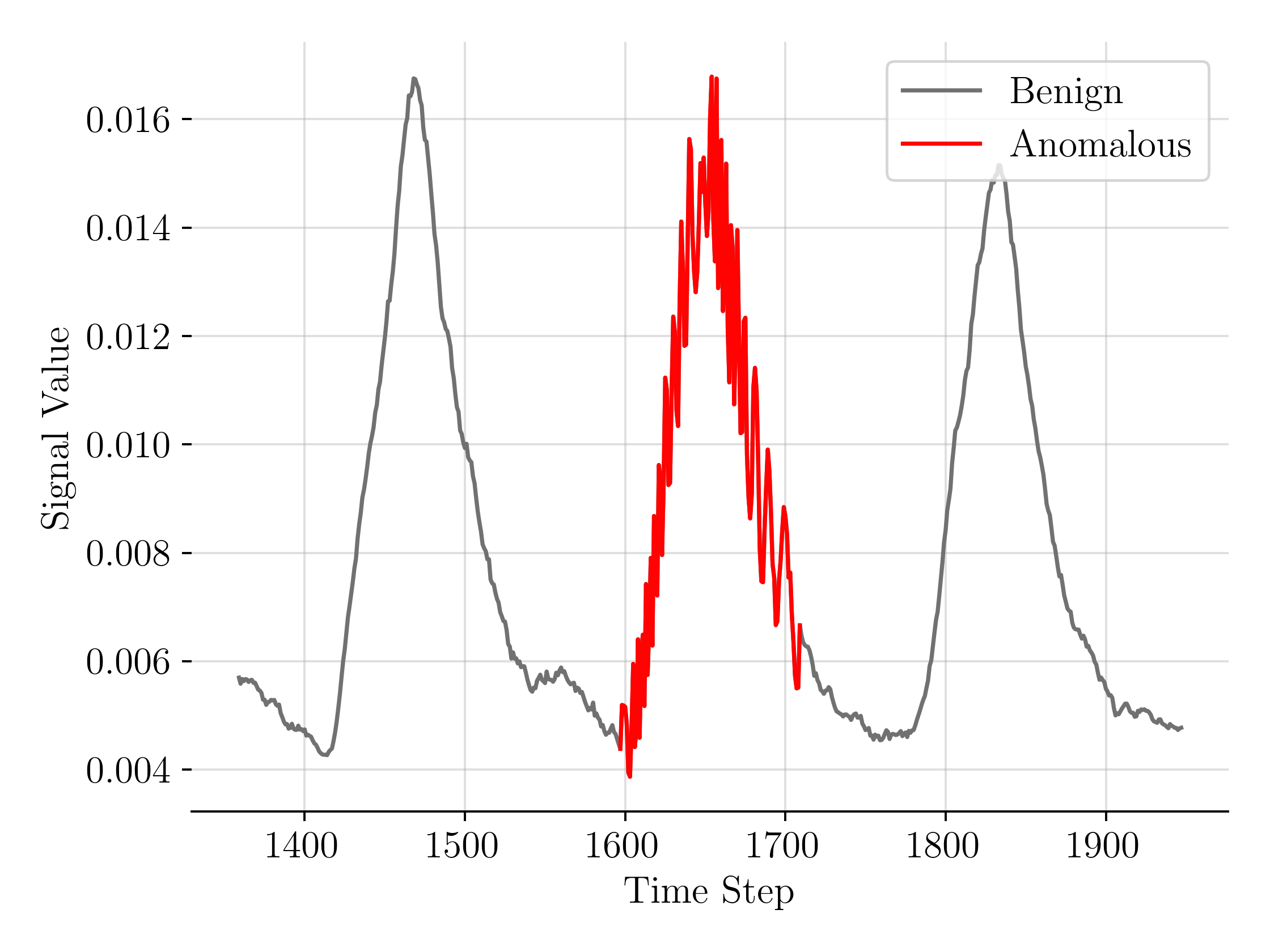}
  \caption{Dataset no. 28}
  \label{fig:DatasetOverviewSfig1}
\end{subfigure}%
\begin{subfigure}{.33\textwidth}
  \centering
  \includegraphics[width=1\linewidth]{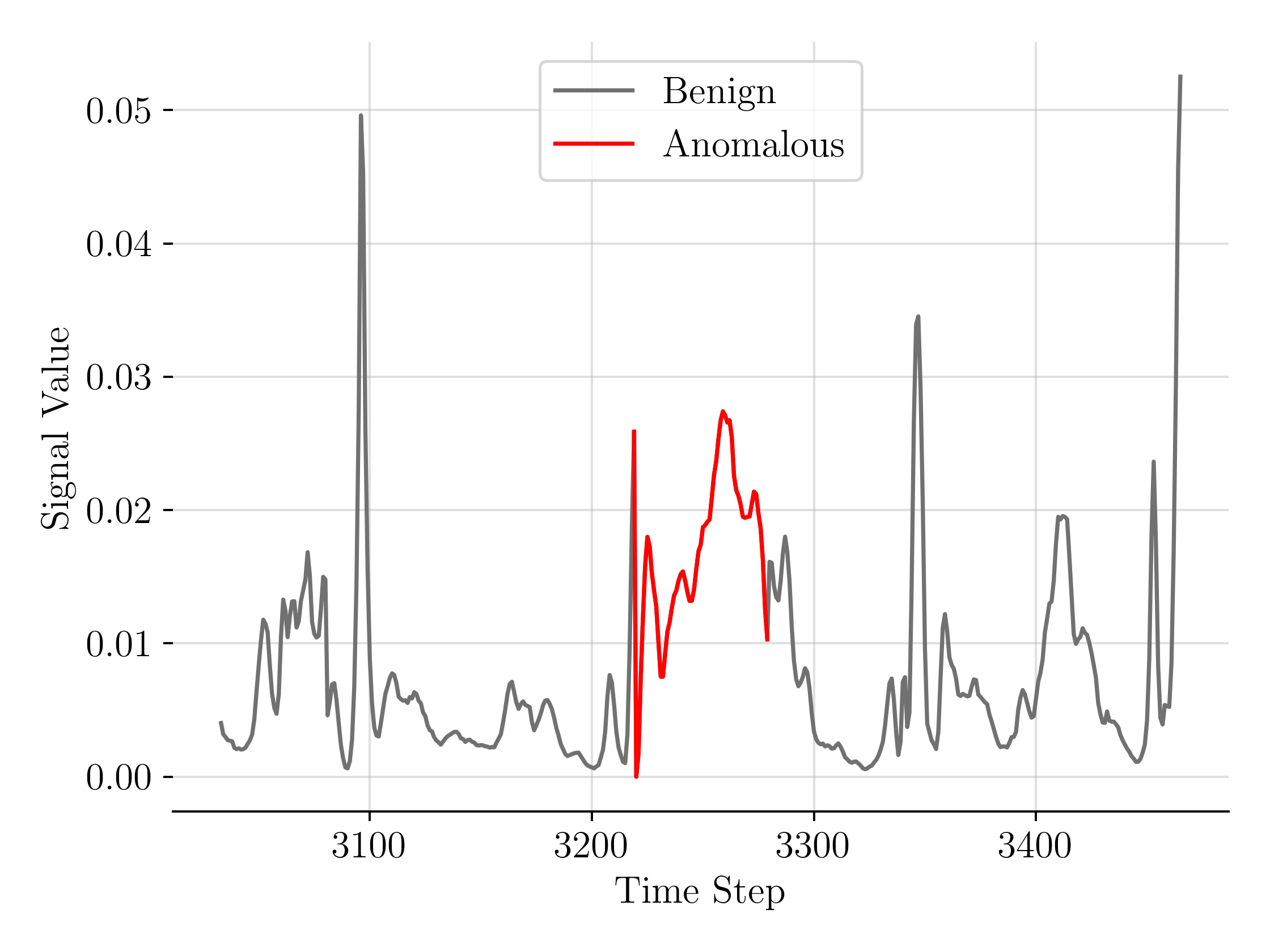}
  \caption{Dataset no. 54}
  \label{fig:DatasetOverviewSfig2}
\end{subfigure}
\begin{subfigure}{.33\textwidth}
  \centering
  \includegraphics[width=1\linewidth]{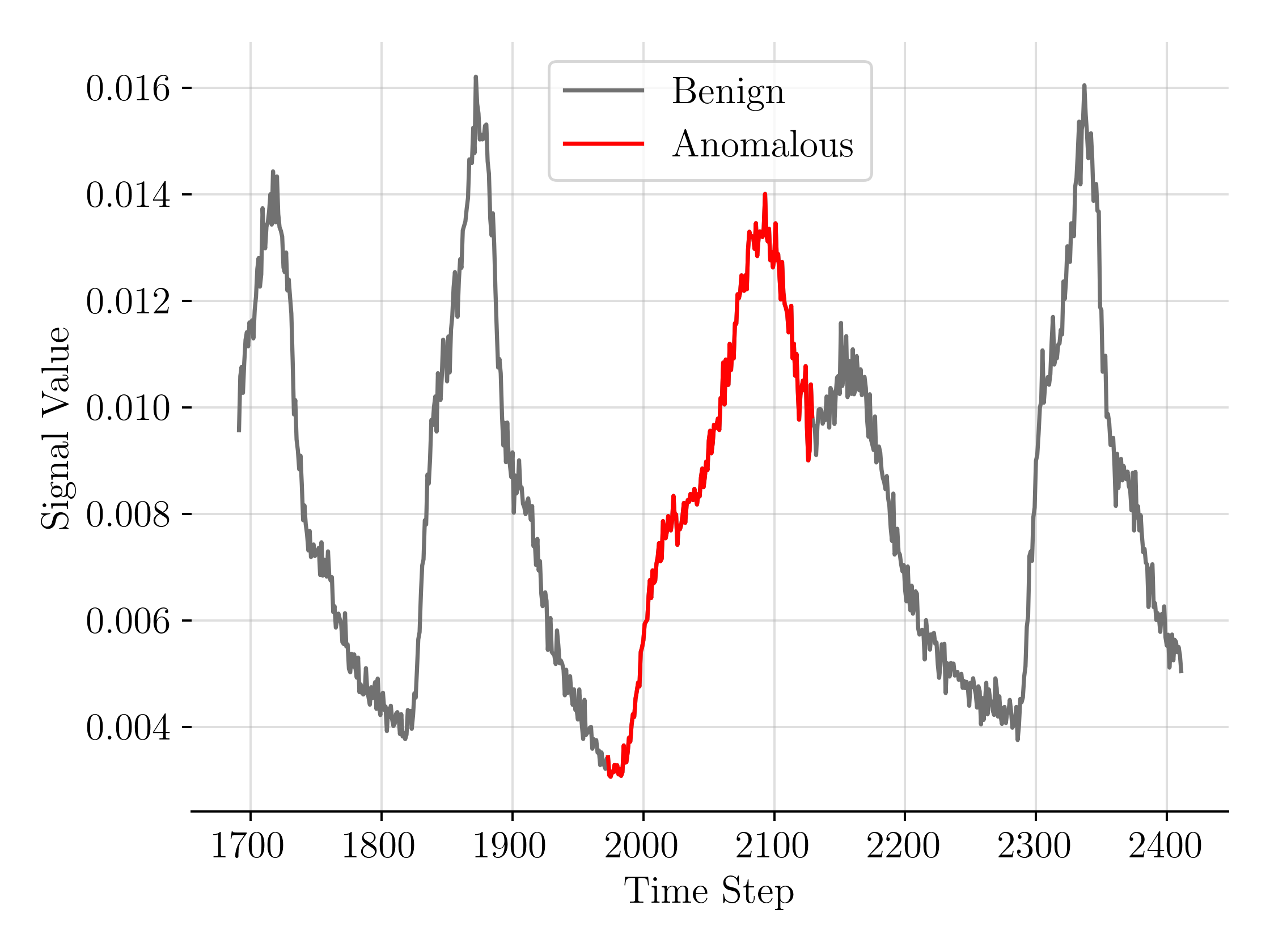}
   \caption{Dataset no. 99}
  \label{fig:DatasetOverviewSfig3}
\end{subfigure}

\begin{subfigure}{.33\textwidth}
  \centering
  \includegraphics[width=1\linewidth]{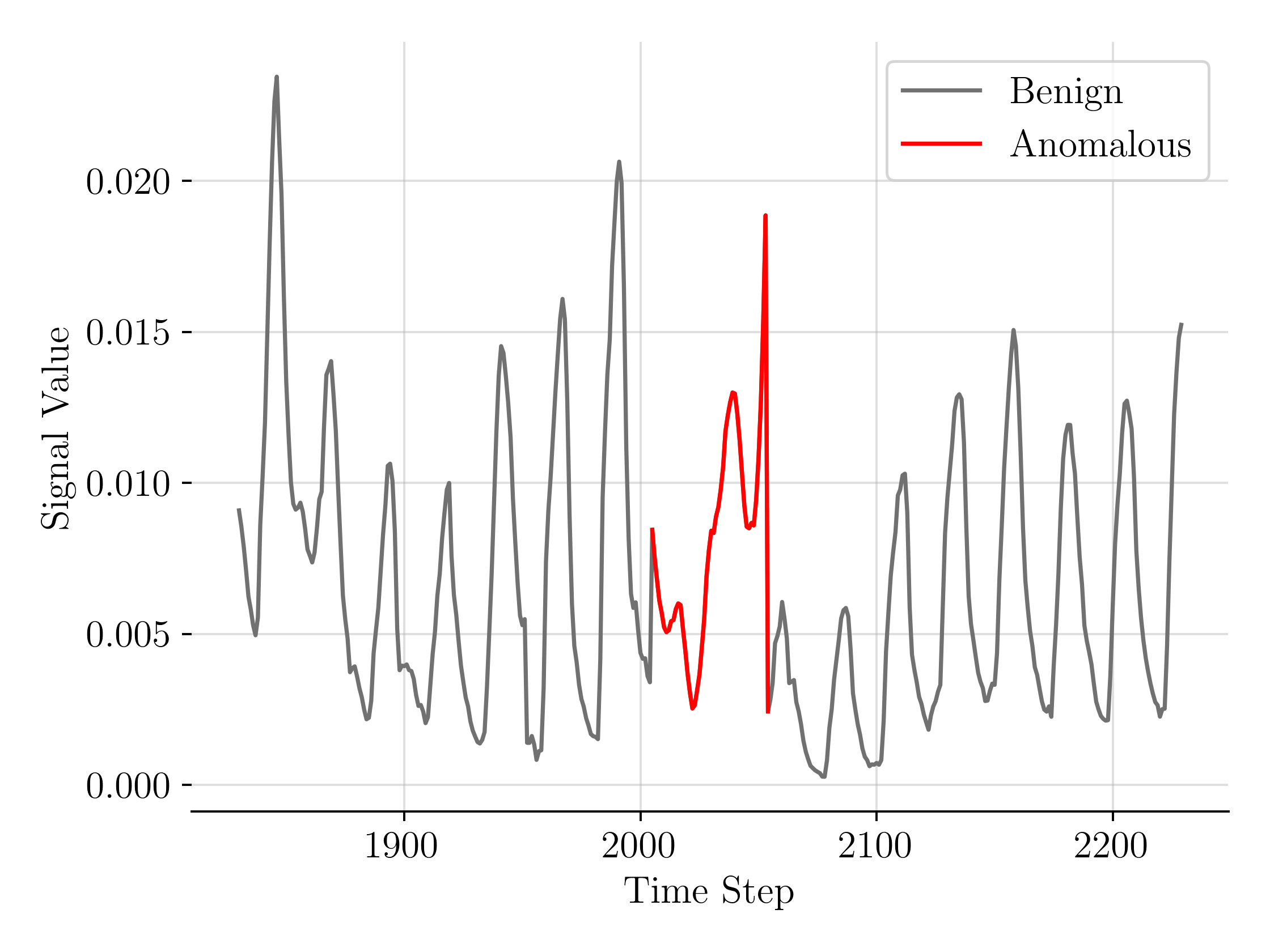}
  \caption{Dataset no. 118}
  \label{fig:DatasetOverviewSfig4}
\end{subfigure}%
\begin{subfigure}{.33\textwidth}
  \centering
  \includegraphics[width=1\linewidth]{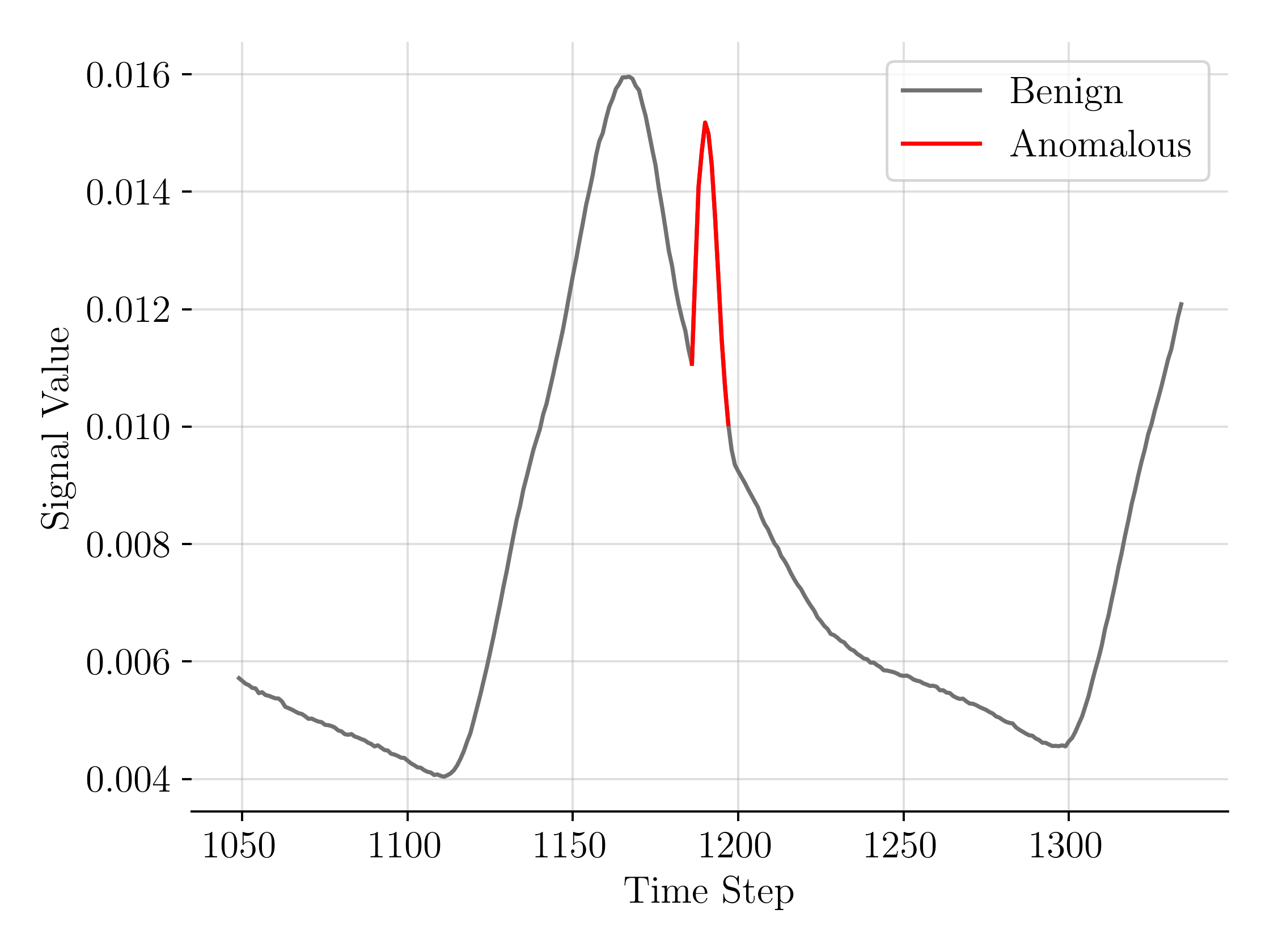}
  \caption{Dataset no. 138}
  \label{fig:DatasetOverviewSfig5}
\end{subfigure}
\begin{subfigure}{.33\textwidth}
  \centering
  \includegraphics[width=1\linewidth]{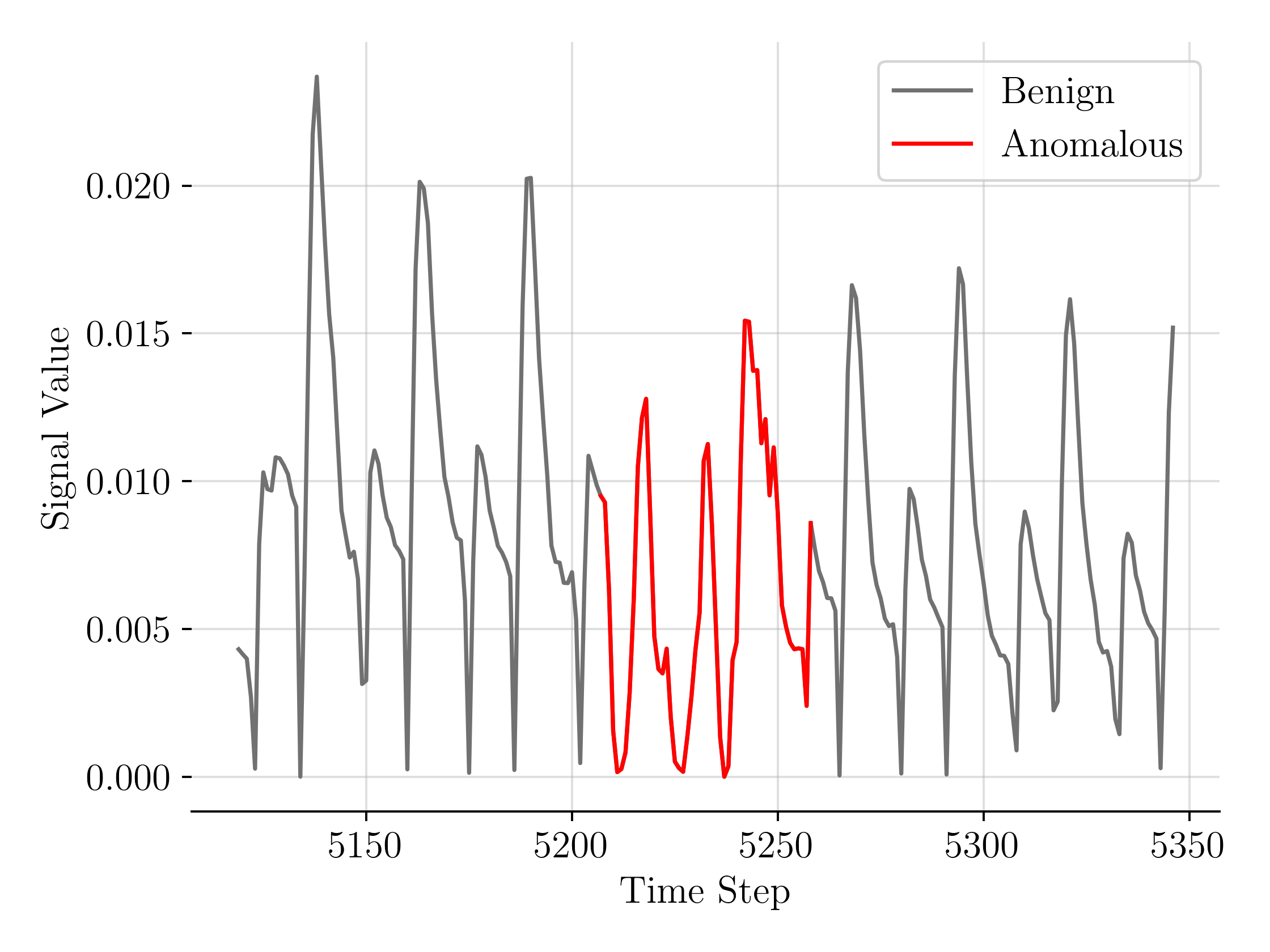}
  \caption{Dataset no. 176}
  \label{fig:DatasetOverviewSfig6}
\end{subfigure}
\caption{Illustration of the datasets employed in this work. The data is an excerpt of the test data with red indicating the anomaly. The anomalies range from relatively easy to spot (dataset 28) to very subtle and hard to spot (dataset 176).}
\label{fig:DatasetOverview}
\end{figure*}

Using a window-based methodology, an anomaly is considered a valid detection of the anomaly if it starts within the valid detection range, which includes both the 'Anomaly Precursor' and 'Anomalous' ranges. The 'Anomaly Precursor' denotes the time windows at time $t$ that precede the anomalous range but overlaps with it.

For our experiments, we focus on a subset of the datasets provided by the authors. An illustration of the datasets used is given in Figure \ref{fig:DatasetOverview}. We preprocess the data into sliding windows of size 128 with a stride of 1 as done by \cite{frehner2024detecting}, with each window serving as input to the autoencoder. 

\subsection{Baseline: Classical Autoencoder}
\label{sect:BaselineSetup}
\begin{figure}[ht]
    \centering
    \includegraphics[width=0.7\linewidth]{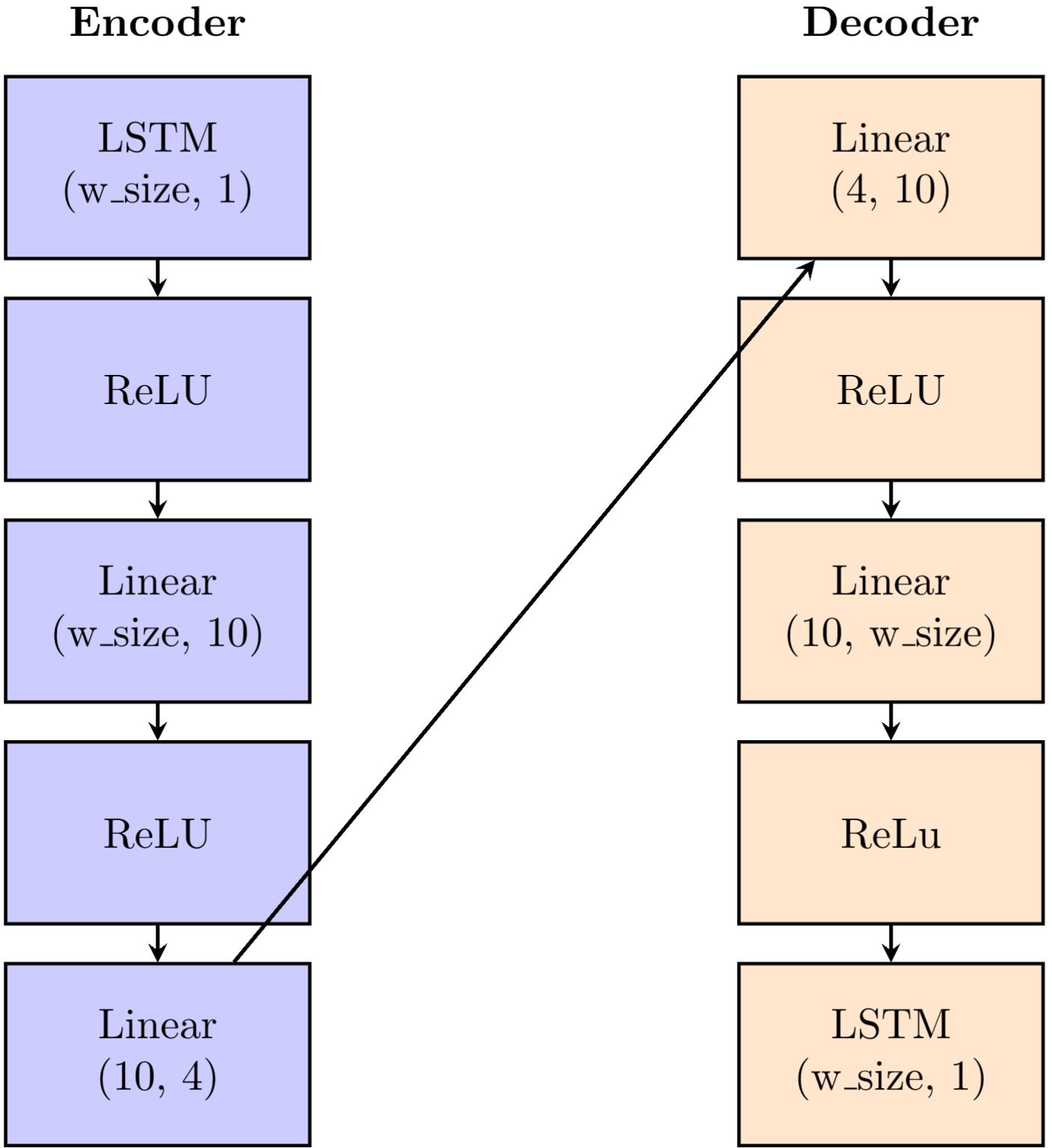}
    \caption{Illustration of the classical deep learning autoencoder used as the baseline in this study. This architecture follows the design employed by \cite{frehner2024detecting}. In this work, the window size is set to 128, while other parameters remain unchanged.}
    \label{fig:BaselineArchitecture}
\end{figure}
To enable a comparative analysis between quantum autoencoders and classical deep learning-based autoencoders, we establish a classical baseline model. The architecture shown in Figure \ref{fig:BaselineArchitecture} replicates the autoencoder utilized by \cite{frehner2024detecting}, which has demonstrated effectiveness on the UCR Benchmark. Given our window size of 128, the input dimension is correspondingly set to 128. The baseline deep learning autoencoder is implemented using the PyTorch framework.


\subsection{Quantum Simulator Experiment}
\label{sect:SimulatorSetup}
For our experiments we use various quantum autoencoder architectures as shown in Table \ref{tab:AnsatzOverview}. Performance will be reported based on the average of 5 executions for each employed ansatz. 

We employ the autoencoder architecture detailed in Section \ref{sect:methodology}, in combination with amplitude encoding for data representation. The input data comprises time windows generated according to the procedure described in Section \ref{sect:UCRDataset}. In our simulated experiments, 7 qubits are allocated for data encoding purposes. During the training phase, elaborated upon in Section \ref{sect:QAEArchitectureTraining}, additional qubits are required, totaling 9 qubits. The latent space consistently utilizes 6 qubits, thereby aiming to compress the initial 7 qubit quantum state into a more concise representation. 
\setlength\tabcolsep{1.5pt}
\begin{table}
\centering
\begin{tabular}{@{}ccrrrrrrrrr@{}}
\toprule
\multicolumn{1}{l}{} & \multicolumn{1}{l}{Ansatz Depth} & 2                   & 3                   & 4                   & 5                   & 6                   & 7                   & 8                   & 9                   & 10                  \\
Ansatz               & Entanglement                     & \multicolumn{9}{c}{No. of Parameters}                                                                                                                                                               \\ \midrule
PauliTwoDesign       & Pairwise CZ                      & \multirow{5}{*}{21} & \multirow{5}{*}{28} & \multirow{5}{*}{35} & \multirow{5}{*}{42} & \multirow{5}{*}{49} & \multirow{5}{*}{56} & \multirow{5}{*}{63} & \multirow{5}{*}{70} & \multirow{5}{*}{77} \\ \cmidrule(r){1-2}
RealAmplitudes        & Circular                         &                     &                     &                     &                     &                     &                     &                     &                     &                     \\ \cmidrule(r){1-2}
RealAmplitudes        & Full                             &                     &                     &                     &                     &                     &                     &                     &                     &                     \\ \cmidrule(r){1-2}
RealAmplitudes        & Linear                           &                     &                     &                     &                     &                     &                     &                     &                     &                     \\ \cmidrule(r){1-2}
RealAmplitudes        & SCA                              &                     &                     &                     &                     &                     &                     &                     &                     &                     \\ \bottomrule
\end{tabular}
\caption{Overview of the ansätze utilized in this study, detailing the entanglement schemes and the corresponding number of trainable parameters. The ansatz depth indicates the number of repetitions of the layered gates for each ansatz.}
\label{tab:AnsatzOverview}
\end{table}
\setlength\tabcolsep{6pt}

\paragraph{RealAmplitudes Ansatz}
The RealAmplitudes ansatz by IBM Qiskit features alternating layers of Y rotations and CX entanglements. The entanglement pattern can be user-defined or chosen from a predefined set. We will use RealAmplitudes with varying \textit{depth} (i.e. the number of repeating Y rotations and entanglement) as well as different entanglement patterns. 

\paragraph{PauliTwoDesign Ansatz}
\label{sect:PauliTwoDesignAnsatz}
This ansatz implements a specific type of circuit, frequently explored in quantum machine learning literature, particularly in studies involving barren plateaus in variational algorithms \cite{McCleanBarrenPlateau}. The circuit comprises alternating rotation and entanglement layers, starting with an initial layer of $R_Y\left(\frac{\pi}{4}\right)$ gates. In the rotation layers, single-qubit Pauli rotations are applied, with the rotation axis randomly chosen from $X$, $Y$, or $Z$. The entanglement layers consist of pairwise $CZ$ gates. For consistency, the same random seed (42) to generate the different ansaetze is used throughout this work.

\paragraph{Quantum Simulator System}
\label{sect:QuantumSimulatorSystem}
For this experiment, the IBM Qiskit 1.0.0 \cite{IBMQiskit} and Qiskit Aer 0.13.1 \cite{IBMQiskit} Python library were used with the statevector as the chosen simulator. No changes were made to the simulator settings and the numbers of shots per execution was set to 5,000 as we only measure 1 qubit without noise simulation. All experiments were conducted on 16 Intel Broadwell CPU cores. For optimization we use COBYLA with default settings and 45 iterations.

\paragraph{Baseline and Anomaly Classification}
The baseline deep learning architecture is trained on the same dataset as the quantum autoencoders, utilizing a batch size of 150, the ADAM optimizer with default settings, and 250 epochs. Anomalies are classified based on the mean squared error between the input and its reconstructed output.

The quantum autoencoders detect anomalies using two approaches: a single Swap Test measurement (architecture shown in Figure \ref{fig:QAEArchitectureSfig1}) and input reconstruction with mean squared error calculation (architecture shown in Figure \ref{fig:QAEArchitectureSfig2}), similar to the baseline method.

Anomaly classification is performed on the post-processed output, following the procedure outlined in Section \ref{sect:MethodologyPostprocessing}. This involves a moving average window of size 128, which corresponds to the time window size. This window size is arbitrarily chosen and is not optimized for each dataset. The time step exhibiting the largest post-processed value is flagged as anomalous.
\begin{figure*}[ht]
    \centering
    \includegraphics[width=1.0\linewidth]{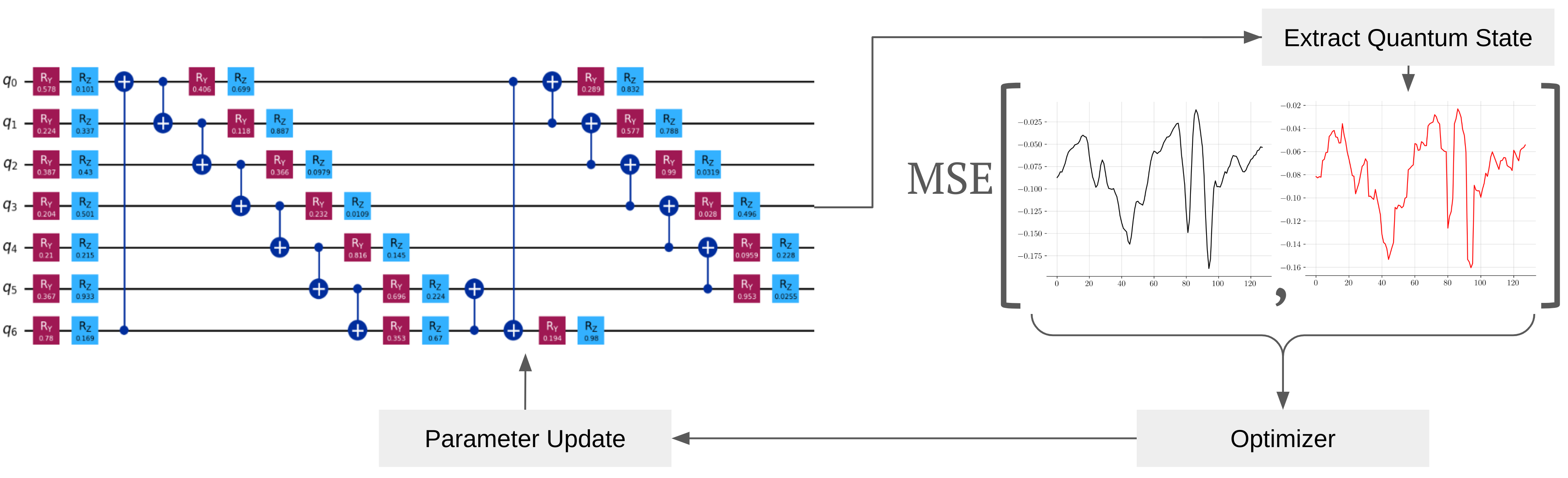}
    \caption{Overview of the quasi-amplitude-encoding procedure utilized in this paper. The method employs a hardware-efficient variational circuit, specifically IBM Qiskit's EfficientSU2 layout with two repetitions using seven qubits. The circuit parameters are optimized iteratively by first extracting the quantum state generated by the current parameter set. The mean squared error (MSE) between this extracted state and the target state is then calculated and provided to the selected optimizer, SPSA in this case, to generate a new parameter set. This optimization loop continues until a specified convergence criterion is met.}
    \label{fig:QuasiEncodingOverview}
\end{figure*}
\subsection{Real Quantum Hardware Experiment}
\label{sect:RealQuantumHardwareExperiment}
For evaluating the quantum autoencoder on real hardware, a couple of changes are required because of the noisiness of NISQ quantum computers. Subsequently we describe the experimental setup for the quantum autoencoder on real hardware. 

\paragraph{Data and Encoding}
For the experiments conducted on a real quantum computer, we will concentrate on a single dataset, specifically dataset no. 54, and further reduce the training data to 150 sliding time windows. The training data spans from $t_s =400$ to $t_e=550+127=677$, thereby comprising 150 time windows and covering 277 time steps. This reduction in data is essential, as each time window corresponds to an individual circuit execution, which incurs substantial quantum computational time.

A critical aspect of executing a quantum circuit on actual hardware is the transpilation process. This involves converting the logical design of the desired circuit into a format suitable for a specific target quantum system, such as IBM Torino \cite{ibmQuantum}. During transpilation, the designed circuit is transformed into a logically equivalent one comprising only the quantum operations supported by the target system. Additionally, at this stage, the logical qubits are mapped onto physical qubits, a process that is both complex and non-trivial. This conversion process can significantly increase the number of operations compared to the original design. 

In the context of amplitude encoding, our findings indicate that the transpiled circuit, optimized for the target system, requires substantially more operations than the initially designed circuit. The excessive number of additional operations renders measurement results unreliable due to quantum state decoherence and the introduction of errors. This substantial increase in quantum operations highlights the deficiencies of amplitude encoding on current quantum hardware. Specifically, the algorithm's inability to efficiently represent amplitude encoding using the basis gates of the target hardware underlines its denotation as hardware inefficient.
\begin{figure}[ht]
    \centering
        \includegraphics[width=1.0\linewidth]{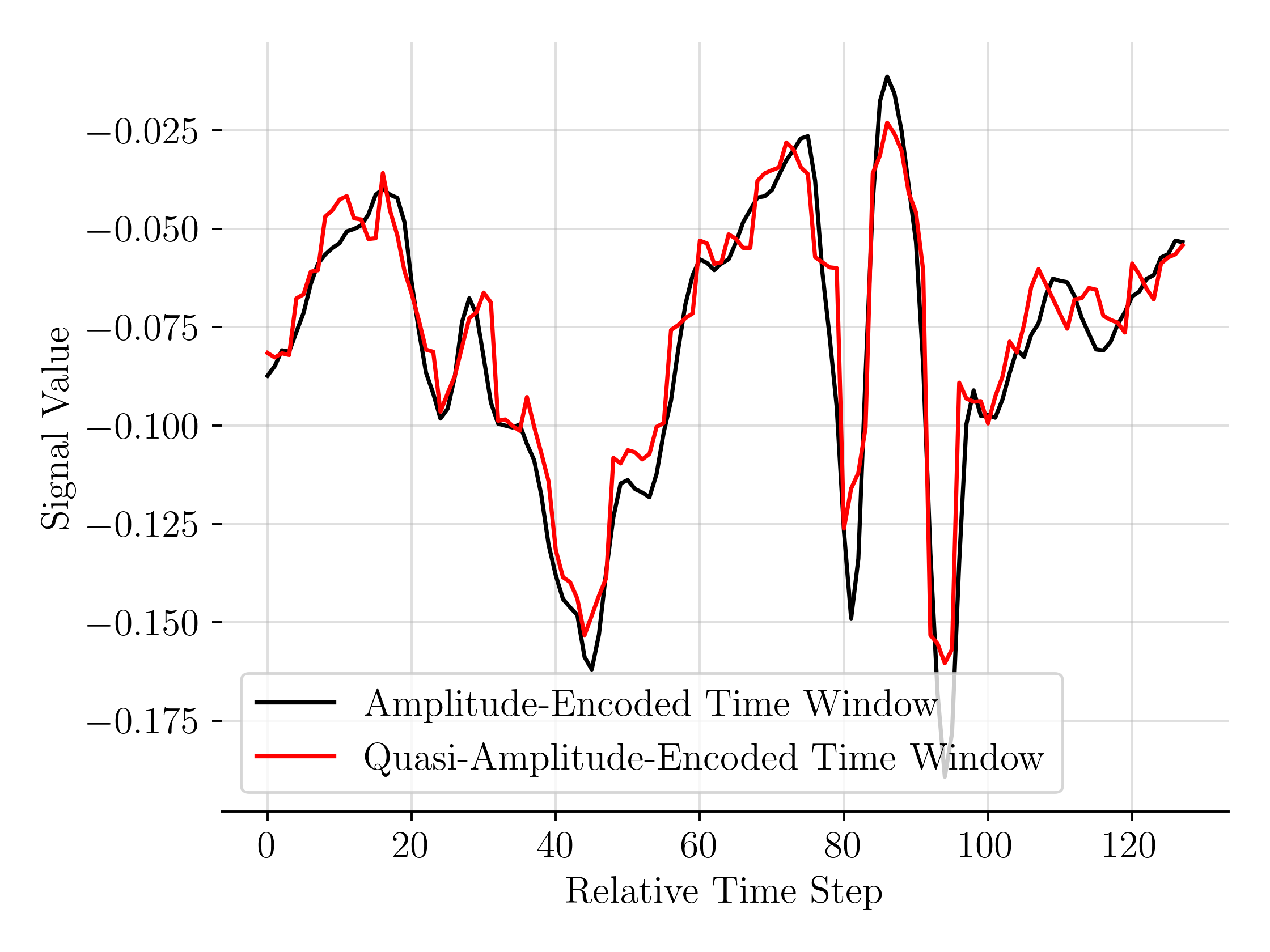}
    \caption{Time series representation of extracted quantum states for a specific time window following amplitude encoding and quasi-amplitude encoding. The depicted time window corresponds to dataset 54, covering time steps 470 to 598. The quasi-amplitude encoded quantum state is derived from the optimal encoding circuit, selected from 100 trials based on the minimization of the mean squared error (MSE) between the target state and the approximated quasi-amplitude encoded state. The amplitude encoded time window exactly matches the input time window.}
    \label{fig:QuasiEncodingExample}
\end{figure}

\paragraph{Quasi Amplitude Encoding}

To address this issue, we employ quasi-amplitude encoding, following the approach outlined in \cite{NakajiQuasiEnc}. Quasi-amplitude encoding is a technique in quantum computing that enables the efficient embedding of classical data into quantum states. Unlike exact amplitude encoding, which requires complex and resource-intensive quantum circuits to precisely represent data values, quasi-amplitude encoding introduces a controlled approximation that significantly reduces circuit complexity while maintaining acceptable accuracy. Figure \ref{fig:QuasiEncodingExample} illustrates a time series comparison between the quantum states produced by amplitude encoding and quasi-amplitude encoding. While amplitude encoding precisely reproduces the original series, quasi-amplitude encoding offers a reasonable approximation of the time window. The overall quasi-amplitude-encoding procedure is depicted in Figure \ref{fig:QuasiEncodingOverview}.

We implement this approach on a classical computer as part of the preprocessing step detailed in Section \ref{sect:MethdologyPreprocessing}. For the quasi-amplitude encoding, we utilize the Qiskit EfficientSU2 layout, as shown in Figure \ref{fig:QuasiEncodingOverview}. The EfficientSU2 layout refers to a parameterized quantum circuit structure optimized for hardware efficiency and versatility in variational quantum algorithms, comprising layers of parameterized single-qubit rotations and CX entanglements, designed to minimize gate count after transpilation.
\begin{figure*}[ht]
    \centering
    \includegraphics[width=1.0\linewidth]{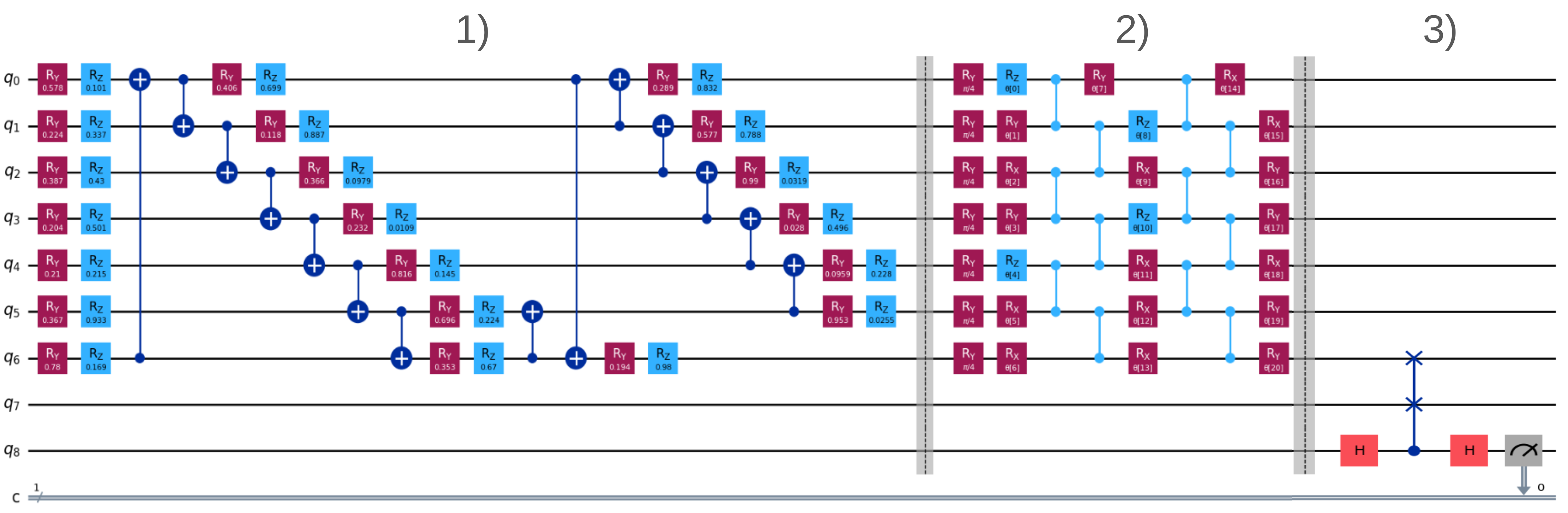}
    \caption{Illustration of a circuit optimized for execution on real quantum hardware before transpilation. Component 1) represents the quasi-encoded time window using the EfficientSU2 circuit layout. The parameter values in this section depend on the input data and remain constant during training. Component 2) illustrates the trainable encoder of the quantum autoencoder, which, in this case, is based on the PauliTwoDesign with two repetitions. These parameters are adjustable and are optimized during training on the actual hardware. Lastly, component 3) corresponds to the Swap-Test which measures how well the encoder circuit encodes the quasi-encoded input data. Ideally the likelihood of measuring qubit 8 in state $1$ is close to 0. This exact circuit is used in this study for execution on real quantum hardware.}
    \label{fig:ExampleRealHardwareCircuit}
\end{figure*}
Each of the 150 time windows in our dataset was encoded into a quantum circuit, and the best result out of 100 runs, based on the mean squared error, was selected. This process yielded 150 circuits (one per time window), each of which can be integrated into the quantum autoencoder architecture by replacing the amplitude encoding component with the respective circuit representing the quasi-amplitude encoded time window. Although quasi-amplitude encoding introduces some noise, the overall shape and characteristics of the time series are preserved, as demonstrated in Figure \ref{fig:QuasiEncodingExample}.

For the test data, we followed a similar procedure, i.e. selecting a subset of the test data that includes the anomalous range and surrounding non-anomalous data. We selected the 100 time windows preceding and 85 succeeding the valid detection range (described in Section \ref{sect:UCRDataset}), as well as the valid detection range itself, encoding them in the same manner. 

The test data covers a range from $t_s = 2993$ to $t_e =3363+127$, encompassing a total of 497 time steps. Since the target is anomaly detection, we need to ensure that the quasi amplitude-encoding process does not introduce additional deviations from the original series that could be mistaken for anomalies. To this end, we employed a Mann-Whitney U test between the residuals of the encoded training data and test data, which resulted in a p-value of 0.09. Although this suggests no significant evidence against the hypothesis that both residual samples originate from the same distribution at the 0.05 significance level, it may impact final performance as the test only holds true on the 5\% level and fails at the 10\%.

For the evaluation on real quantum hardware, we restrict our analysis to a single quantum autoencoder configuration using the described encoded time windows and a PauliTwo ansatz with 2 repetitions, totaling 21 trainable parameters on the quantum computer. This parameter count is determined by the selected ansatz.

\paragraph{Further Hardware Optimization}
To further optimize for hardware efficiency, we narrow our focus to anomaly identification using the quantum autoencoder architecture depicted in Figure \ref{fig:QAEArchitectureSfig1} and classify anomalies based on Swap-Test measurements. This method reduces the number of operations required by omitting the decoding step and solely evaluating the encoder's efficacy in representing the data in the latent space. This evaluation is consistent with the measurement minimized during the training phase.

The training process involves the 150 training time windows over 45 epochs, utilizing COBYLA with default parameters for optimization and 8192 shots. We chose 8192 as it is the maximum available on the IBM Torino machine at the time of writing and to ensure best possible error mitigation. All experiments on real quantum hardware are conducted on the 133-qubit IBM Torino system in session mode, leveraging the Sampler primitive \cite{QiskitSampler}. The Sampler primitive generates a quasi-probability distribution with built-in error mitigation, in contrast to raw measurement counts. In this work, we leverage this integrated error mitigation. 

The final circuit design employed in this paper is illustrated in Figure \ref{fig:ExampleRealHardwareCircuit}. This circuit will be utilized for experiments on real quantum hardware, with Component 1 being substituted by the quasi-encoding circuit corresponding to the specific input data being encoded.
\begin{table*}[ht]
\resizebox{1\linewidth}{!}{
\begin{tabular}{@{}clrrrrrrrrrlrrrrrrrrr@{}}
\toprule
\multicolumn{1}{l}{}                                  & \multicolumn{1}{c}{}     & \multicolumn{9}{c}{MSE-based Detection}                                                                                                       & \multicolumn{1}{c}{} & \multicolumn{9}{c}{Swap-based Detection}                                                                                                      \\ \midrule
\multicolumn{1}{l}{}                                  & Number of Parameters     & 21            & 28            & 35            & 42            & 49            & 56            & 63            & 70            & 77            &                      & 21            & 28            & 35            & 42            & 49            & 56            & 63            & 70            & 77            \\
\begin{tabular}[c]{@{}c@{}}Dataset\\ no.\end{tabular} & Ansatz                   & \multicolumn{9}{r}{}                                                                                                                          &                      & \multicolumn{9}{r}{}                                                                                                                          \\ \midrule
\multirow{6}{*}{28}                                   & PauliTwoDesign           & 0.00          & 0.00          & 0.00          & 0.00          & 0.00          & 0.00          & 0.00          & 0.00          & 0.00          &                      & 0.00          & 0.00          & 0.00          & 0.00          & 0.00          & 0.00          & 0.00          & 0.00          & 0.00          \\
                                                      & RealAmplitude - Circular & 0.00          & 0.00          & 0.00          & 0.00          & 0.00          & 0.00          & 0.00          & 0.00          & 0.00          &                      & 0.00          & 0.00          & 0.00          & 0.00          & 0.00          & 0.00          & 0.00          & 0.00          & 0.00          \\
                                                      & RealAmplitude - Full     & 0.00          & 0.00          & 0.00          & 0.00          & 0.00          & 0.00          & 0.00          & 0.00          & 0.00          &                      & 0.00          & 0.00          & 0.00          & 0.00          & 0.00          & 0.00          & 0.00          & 0.00          & 0.00          \\
                                                      & RealAmplitude - Linear   & 0.00          & 0.00          & 0.00          & 0.00          & 0.00          & 0.00          & 0.00          & 0.00          & 0.00          &                      & 0.00          & 0.00          & 0.00          & 0.00          & 0.00          & 0.00          & 0.00          & 0.00          & 0.00          \\
                                                      & RealAmplitude - SCA      & 0.00          & 0.00          & 0.20          & 0.00          & 0.00          & 0.00          & 0.20          & 0.00          & 0.00          &                      & 0.00          & 0.00          & \textbf{0.40} & 0.00          & \textbf{0.40} & 0.00          & 0.20          & 0.00          & 0.00          \\
                                                      & Classical Autoencoder     & \multicolumn{9}{c}{0.20}                                                                                                                      &                      & \multicolumn{9}{c}{0.20}                                                                                                                      \\ \cmidrule(r){1-11} \cmidrule(l){13-21} 
\multicolumn{11}{c}{}                                                                                                                                                                                                            &                      & \multicolumn{9}{l}{}                                                                                                                          \\ \cmidrule(r){1-11} \cmidrule(l){13-21} 
\multirow{6}{*}{54}                                   & PauliTwoDesign           & \textbf{1.00} & \textbf{1.00} & \textbf{1.00} & \textbf{1.00} & \textbf{1.00} & \textbf{0.80} & \textbf{0.80} & \textbf{0.80} & 0.20          &                      & \textbf{1.00} & \textbf{1.00} & \textbf{1.00} & \textbf{1.00} & \textbf{1.00} & \textbf{1.00} & \textbf{1.00} & \textbf{1.00} & \textbf{1.00} \\
                                                      & RealAmplitude - Circular & 0.20          & 0.00          & 0.20          & 0.00          & 0.00          & 0.00          & 0.00          & 0.00          & 0.00          &                      & 0.20          & 0.00          & 0.20          & 0.00          & 0.20          & 0.00          & 0.00          & 0.00          & 0.00          \\
                                                      & RealAmplitude - Full     & 0.00          & 0.00          & 0.20          & 0.00          & 0.20          & 0.00          & 0.00          & 0.00          & 0.00          &                      & \textbf{1.00} & 0.00          & 0.20          & \textbf{0.40} & \textbf{0.40} & 0.20          & \textbf{0.60} & \textbf{0.40} & \textbf{0.60} \\
                                                      & RealAmplitude - Linear   & \textbf{0.80} & \textbf{0.80} & \textbf{0.60} & \textbf{0.60} & 0.20          & 0.00          & 0.00          & 0.00          & 0.00          &                      & \textbf{1.00} & \textbf{0.80} & \textbf{0.80} & \textbf{0.60} & 0.20          & \textbf{0.40} & \textbf{0.40} & 0.00          & 0.20          \\
                                                      & RealAmplitude - SCA      & \textbf{0.80} & 0.20          & 0.00          & 0.20          & 0.00          & 0.00          & 0.00          & 0.00          & 0.00          &                      & \textbf{0.80} & \textbf{0.60} & 0.00          & \textbf{0.60} & 0.20          & 0.00          & 0.00          & 0.00          & 0.00          \\
                                                      & Classical Autoencoder     & \multicolumn{9}{c}{0.20}                                                                                                                      & \multicolumn{1}{c}{} & \multicolumn{9}{c}{0.20}                                                                                                                      \\ \cmidrule(r){1-11} \cmidrule(l){13-21} 
\multicolumn{11}{c}{}                                                                                                                                                                                                            &                      & \multicolumn{9}{l}{}                                                                                                                          \\ \cmidrule(r){1-11} \cmidrule(l){13-21} 
\multirow{6}{*}{99}                                   & PauliTwoDesign           & 0.00          & 0.00          & 0.00          & 0.00          & 0.00          & \textbf{0.40} & 0.00          & 0.00          & 0.00          &                      & 0.00          & 0.00          & 0.20          & 0.00          & 0.00          & \textbf{0.40} & 0.00          & 0.00          & 0.00          \\
                                                      & RealAmplitude - Circular & 0.00          & 0.00          & 0.00          & 0.20          & 0.00          & 0.00          & 0.00          & 0.00          & 0.20          &                      & 0.00          & 0.00          & 0.00          & 0.20          & 0.00          & 0.00          & 0.00          & 0.20          & \textbf{0.40} \\
                                                      & RealAmplitude - Full     & 0.00          & 0.00          & 0.00          & 0.20          & 0.20          & 0.00          & 0.20          & 0.00          & \textbf{0.40} &                      & 0.20          & 0.00          & 0.00          & 0.20          & 0.20          & 0.00          & 0.00          & 0.20          & 0.20          \\
                                                      & RealAmplitude - Linear   & 0.00          & 0.20          & 0.00          & 0.00          & 0.20          & 0.20          & 0.20          & 0.00          & 0.20          &                      & 0.00          & 0.00          & 0.00          & 0.00          & 0.20          & 0.20          & \textbf{0.40} & 0.20          & 0.20          \\
                                                      & RealAmplitude - SCA      & 0.00          & 0.00          & 0.00          & 0.00          & 0.20          & 0.00          & 0.00          & \textbf{0.40} & 0.20          &                      & 0.00          & 0.00          & 0.00          & 0.00          & 0.00          & 0.00          & 0.00          & 0.00          & 0.20          \\
                                                      & Classical Autoencoder     & \multicolumn{9}{c}{0.20}                                                                                                                      & \multicolumn{1}{c}{} & \multicolumn{9}{c}{0.20}                                                                                                                      \\ \cmidrule(r){1-11} \cmidrule(l){13-21} 
\multicolumn{11}{c}{}                                                                                                                                                                                                            &                      & \multicolumn{9}{l}{}                                                                                                                          \\ \cmidrule(r){1-11} \cmidrule(l){13-21} 
\multirow{6}{*}{118}                                  & PauliTwoDesign           & 0.00          & 0.00          & 0.00          & 0.00          & 0.00          & 0.20          & 0.00          & \textbf{0.20} & 0.00          &                      & \textbf{0.40} & \textbf{0.40} & \textbf{0.40} & \textbf{0.60} & \textbf{0.60} & \textbf{0.60} & 0.20          & \textbf{1.00} & \textbf{0.60} \\
                                                      & RealAmplitude - Circular & 0.20          & \textbf{0.40} & 0.00          & 0.20          & \textbf{0.40} & 0.00          & 0.20          & 0.00          & 0.20          &                      & \textbf{0.80} & \textbf{0.80} & \textbf{0.80} & \textbf{0.60} & \textbf{1.00} & \textbf{1.00} & \textbf{0.80} & \textbf{0.60} & \textbf{1.00} \\
                                                      & RealAmplitude - Full     & 0.00          & \textbf{0.80} & \textbf{0.60} & 0.20          & \textbf{0.40} & \textbf{0.40} & 0.20          & \textbf{0.40} & 0.20          &                      & \textbf{0.80} & \textbf{1.00} & \textbf{1.00} & \textbf{0.80} & \textbf{0.40} & \textbf{0.80} & \textbf{0.60} & \textbf{0.80} & \textbf{0.80} \\
                                                      & RealAmplitude - Linear   & 0.00          & 0.00          & 0.00          & 0.20          & 0.40          & 0.20          & \textbf{0.80} & 0.20          & 0.20          &                      & 0.20          & \textbf{0.60} & 0.20          & \textbf{0.60} & \textbf{0.80} & \textbf{0.60} & \textbf{0.80} & \textbf{0.80} & \textbf{0.80} \\
                                                      & RealAmplitude - SCA      & 0.00          & 0.20          & \textbf{0.60} & 0.20          & 0.00          & 0.20          & 0.20          & 0.00          & 0.00          &                      & \textbf{0.60} & \textbf{0.60} & \textbf{0.60} & \textbf{0.80} & 0.00          & \textbf{0.80} & \textbf{0.80} & \textbf{0.60} & \textbf{1.00} \\
                                                      & Classical Autoencoder     & \multicolumn{9}{c}{0.20}                                                                                                                      & \multicolumn{1}{c}{} & \multicolumn{9}{c}{0.20}                                                                                                                      \\ \cmidrule(r){1-11} \cmidrule(l){13-21} 
\multicolumn{11}{c}{}                                                                                                                                                                                                            &                      & \multicolumn{9}{l}{}                                                                                                                          \\ \cmidrule(r){1-11} \cmidrule(l){13-21} 
\multirow{6}{*}{138}                                  & PauliTwoDesign           & 0.00          & 0.00          & 0.00          & \textbf{0.40} & \textbf{0.40} & 0.20          & \textbf{0.40} & 0.00          & 0.00          &                      & 0.20          & 0.00          & 0.00          & \textbf{0.40} & \textbf{0.80} & 0.20          & 0.00          & 0.20          & 0.00          \\
                                                      & RealAmplitude - Circular & 0.00          & 0.00          & 0.00          & 0.00          & 0.00          & 0.00          & 0.00          & 0.00          & 0.00          &                      & 0.00          & 0.00          & 0.20          & 0.00          & \textbf{0.40} & 0.20          & 0.00          & 0.00          & 0.20          \\
                                                      & RealAmplitude - Full     & 0.00          & 0.00          & 0.00          & 0.20          & 0.20          & 0.00          & 0.00          & 0.00          & 0.20          &                      & 0.20          & 0.00          & 0.00          & \textbf{0.40} & 0.20          & 0.00          & 0.20          & 0.00          & 0.20          \\
                                                      & RealAmplitude - Linear   & 0.00          & 0.20          & 0.00          & 0.00          & 0.00          & 0.00          & 0.00          & 0.00          & 0.20          &                      & 0.20          & 0.20          & 0.00          & 0.00          & 0.00          & \textbf{0.40} & 0.20          & 0.20          & 0.00          \\
                                                      & RealAmplitude - SCA      & 0.00          & 0.00          & \textbf{0.60} & 0.00          & 0.20          & 0.20          & \textbf{0.40} & 0.00          & 0.00          &                      & \textbf{0.40} & 0.00          & \textbf{0.60} & 0.20          & \textbf{0.60} & \textbf{0.40} & \textbf{0.60} & 0.20          & \textbf{0.40} \\
                                                      & Classical Autoencoder     & \multicolumn{9}{c}{0.20}                                                                                                                      & \multicolumn{1}{c}{} & \multicolumn{9}{c}{0.20}                                                                                                                      \\ \cmidrule(r){1-11} \cmidrule(l){13-21} 
\multicolumn{11}{c}{}                                                                                                                                                                                                            &                      & \multicolumn{9}{l}{}                                                                                                                          \\ \cmidrule(r){1-11} \cmidrule(l){13-21} 
\multirow{6}{*}{176}                                  & PauliTwoDesign           & 0.00          & 0.00          & 0.00          & \textbf{0.20} & 0.00          & 0.00          & 0.00          & 0.00          & 0.00          &                      & \textbf{1.00} & \textbf{1.00} & \textbf{1.00} & \textbf{1.00} & \textbf{1.00} & \textbf{0.80} & \textbf{1.00} & \textbf{0.80} & \textbf{1.00} \\
                                                      & RealAmplitude - Circular & 0.00          & 0.00          & \textbf{0.20} & \textbf{0.20} & 0.00          & 0.00          & 0.00          & \textbf{0.20} & 0.00          &                      & \textbf{0.60} & \textbf{0.80} & \textbf{1.00} & \textbf{0.80} & \textbf{0.80} & \textbf{0.80} & \textbf{0.60} & \textbf{1.00} & \textbf{0.40} \\
                                                      & RealAmplitude - Full     & 0.00          & 0.00          & 0.00          & 0.00          & \textbf{0.20} & \textbf{0.40} & 0.00          & \textbf{0.40} & 0.00          &                      & \textbf{0.40} & \textbf{0.80} & \textbf{0.20} & \textbf{0.60} & \textbf{0.80} & \textbf{1.00} & \textbf{0.80} & \textbf{0.20} & \textbf{0.80} \\
                                                      & RealAmplitude - Linear   & \textbf{0.60} & \textbf{0.80} & \textbf{0.60} & 0.00          & \textbf{0.40} & \textbf{0.20} & 0.00          & \textbf{0.20} & \textbf{0.20} &                      & \textbf{1.00} & \textbf{0.40} & \textbf{0.80} & \textbf{1.00} & \textbf{1.00} & \textbf{1.00} & \textbf{1.00} & \textbf{1.00} & \textbf{0.80} \\
                                                      & RealAmplitude - SCA      & 0.00          & \textbf{0.40} & \textbf{0.60} & 0.00          & \textbf{0.20} & 0.00          & 0.00          & 0.00          & \textbf{0.20} &                      & \textbf{1.00} & \textbf{0.60} & \textbf{0.00} & \textbf{0.60} & \textbf{0.40} & \textbf{0.80} & \textbf{0.60} & \textbf{0.80} & \textbf{0.80} \\
                                                      & Classical Autoencoder     & \multicolumn{9}{c}{0.00}                                                                                                                      & \multicolumn{1}{c}{} & \multicolumn{9}{c}{0.00}                                                                                                                      \\ \bottomrule
\end{tabular}
}
\caption{Overview of anomaly detection performance for the different ansätze employed in this work on the selected datasets with higher values indicating better performance. On the left hand side, anomaly detection is done using the mean squared error between input and reconstructed output, corresponding to circuits following the architecture depicted in Figure \ref{fig:QAEArchitectureSfig2}. On the right hand side, anomaly detection is based on the Swap-Test measurement after encoding. This approach does not rely on reconstructing the original input but measures how well the data can be encoded and corresponds to the circuit layout depicted in Figure \ref{fig:QAEArchitectureSfig1} where the reconstruction part is omitted (i.e the qubit reset and inverse transform). The bold numbers indicate superior performance over the classical autoencoder. The underlying quantum autoencoders are shared and only the detection approach differs between the two methodologies (i.e the same quantum autoencoder is evaluated using both methodologies). Detection rates for the quantum autoencoders as well as for the classical baseline are averaged over 5 executions.}
\label{tab:AnomalyDetectionPerformance}
\end{table*}

\paragraph{Baseline Setup and Anomaly Classification}
\label{sect:BaselineAndClassificationSetup}
To enable meaningful comparison, the baseline deep learning-based autoencoder uses the quasi-amplitude encoded time windows to ensure that any additional complexity introduced by quasi amplitude-encoding the time windows does not disadvantage the quantum autoencoder. Consequently, both the training and test data consist of the extracted quantum states from the quasi amplitude-encoded circuits corresponding to the associated time windows. 

\section{Results}
\label{sect:results}

In this section we discuss the results of our anomaly detection experiments on a quantum simulator as well as on a real quantum device.


\subsection{Results on Quantum Simulator}
\label{sect:SimulatorResults}

\begin{figure*}
\centering
\begin{subfigure}{.5\textwidth}
  \centering
  \includegraphics[width=1\linewidth]{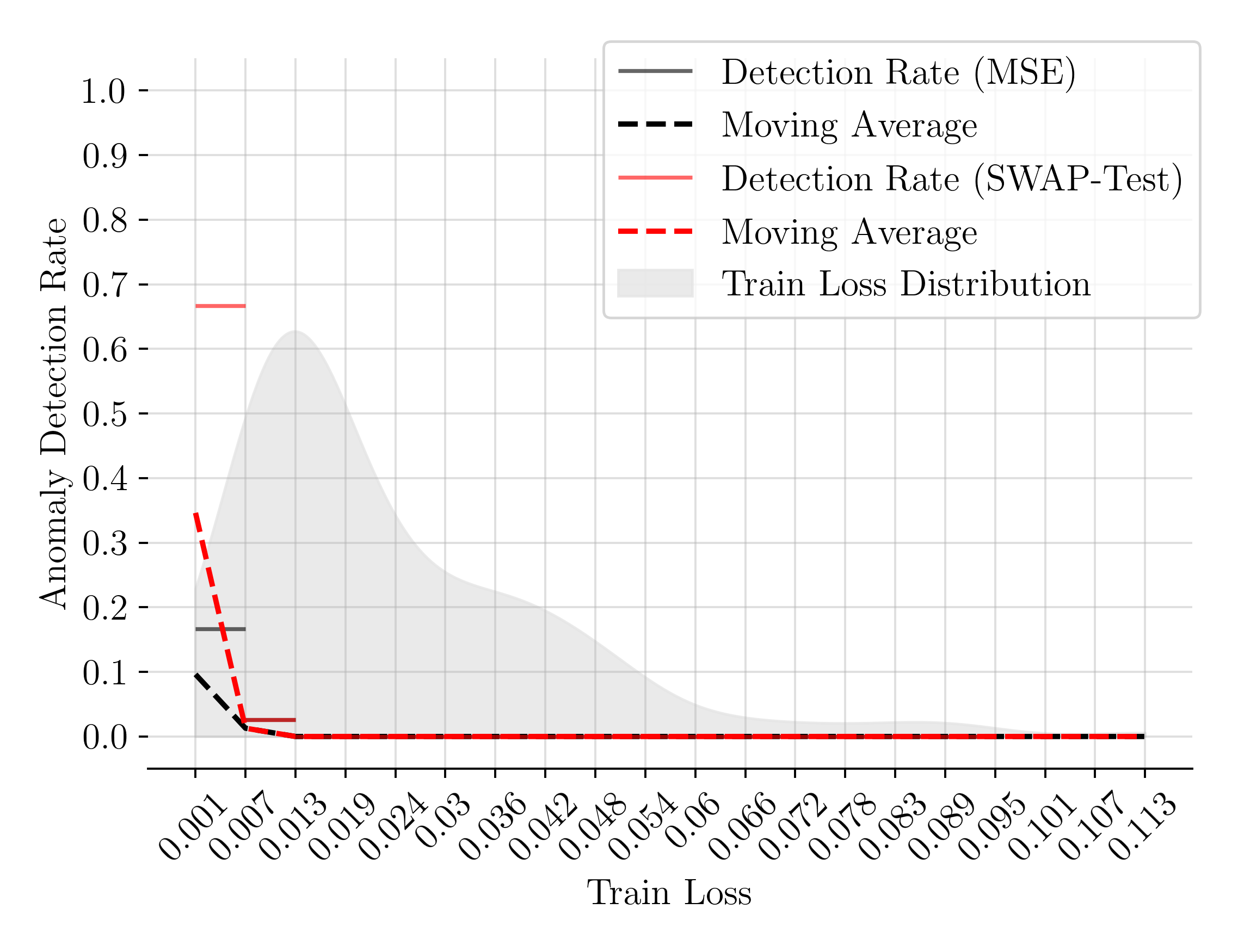}
  \caption{Anomaly detection performance }
  \label{fig:Set28Sfig1}
\end{subfigure}%
\begin{subfigure}{.5\textwidth}
  \centering
  \includegraphics[width=1\linewidth]{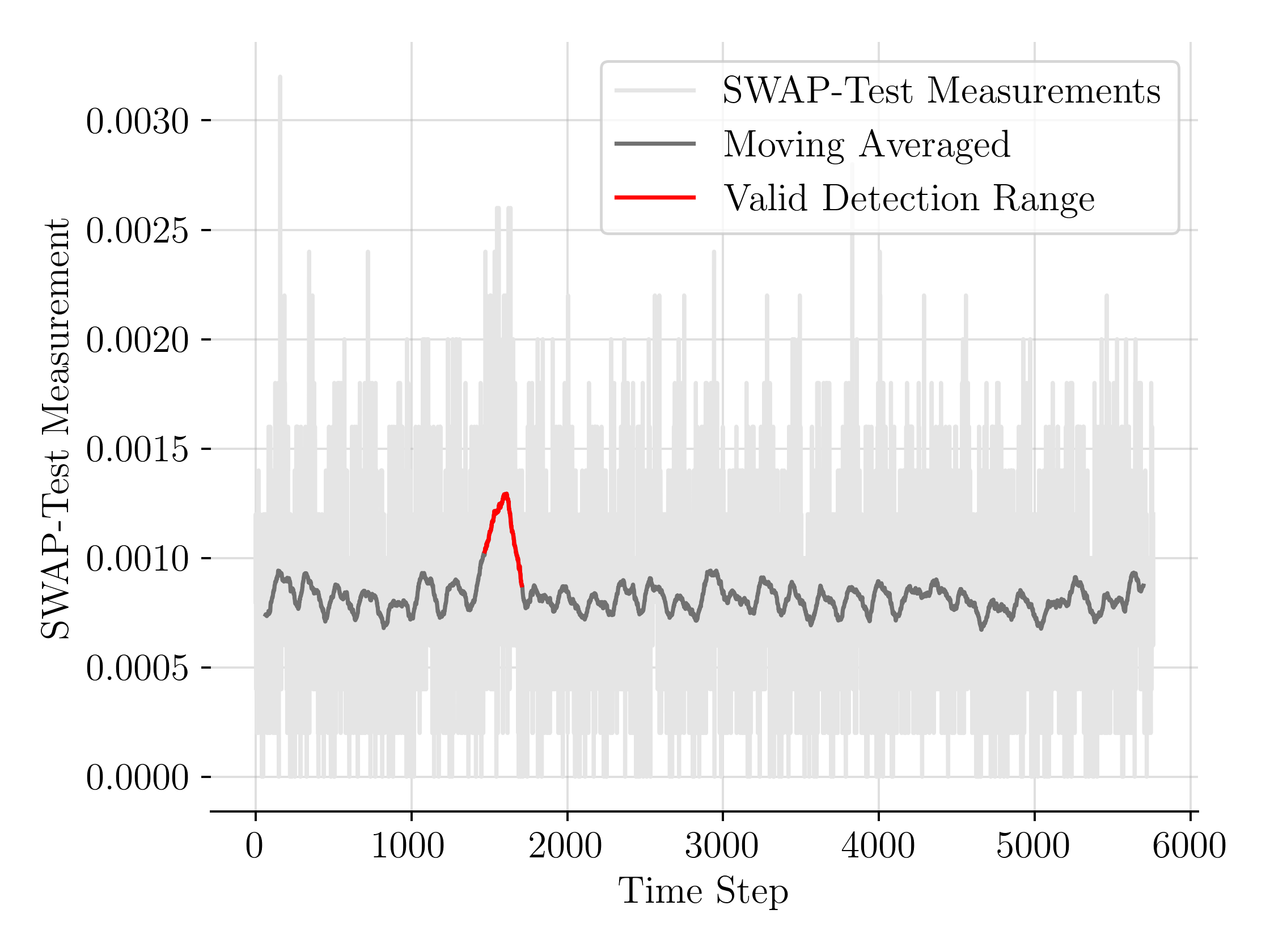}
  \caption{Swap-Test measurements}
  \label{fig:Set28Sfig2}
\end{subfigure}
\caption{Subfigure \ref{fig:Set28Sfig1} depicts the anomaly detection rate for ansaetze on dataset no. 28 within specified train loss intervals, comparing the MSE-based and Swap-Test-based classification methodologies. The detection rate represents the average number of correctly identified anomalies for ansaetze within each train loss interval, with higher values indicating better performance. For example, the Swap-based classification approach accurately identified approximately $\frac{2}{3}$ of the ansaetze with train losses in the interval $[0.001,0.007]$. The density indicates where the different ansaetze converged to. Subfigure \ref{fig:Set28Sfig2} displays the Swap measurements for the ansatz with the lowest train loss observed, including a moving average with a window size of 128 and highlighting the valid detection range. The spike in Swap-Test measurements within the valid detection range is desirable, as anomalies are expected to result in higher Swap measurements and the anomaly can be correctly identified in this case.}
\label{fig:Set28}
\end{figure*}

Table \ref{tab:AnomalyDetectionPerformance} displays the performance of various tested ansaetze, categorized by dataset and the number of trainable parameters for each ansatz. The left section highlights the performance in anomaly identification based on the post-processed Mean Squared Error (MSE) between the input time window and the reconstructed, measured output. On the right, the table shows performance metrics for anomaly identification using post-processed Swap-Test measurements (i.e. no reconstruction of the input). These values are derived from the average of five executions for each combination of dataset, ansatz, and number of parameters. Values near 0 indicate an inability to correctly label anomalies, whereas values close to 1 suggest that the configuration successfully identified anomalies in every execution. The numbers in bold indicate superior performance over the classical autoencoder.

\begin{figure}[ht]
    \centering
    \includegraphics[width=1\linewidth]{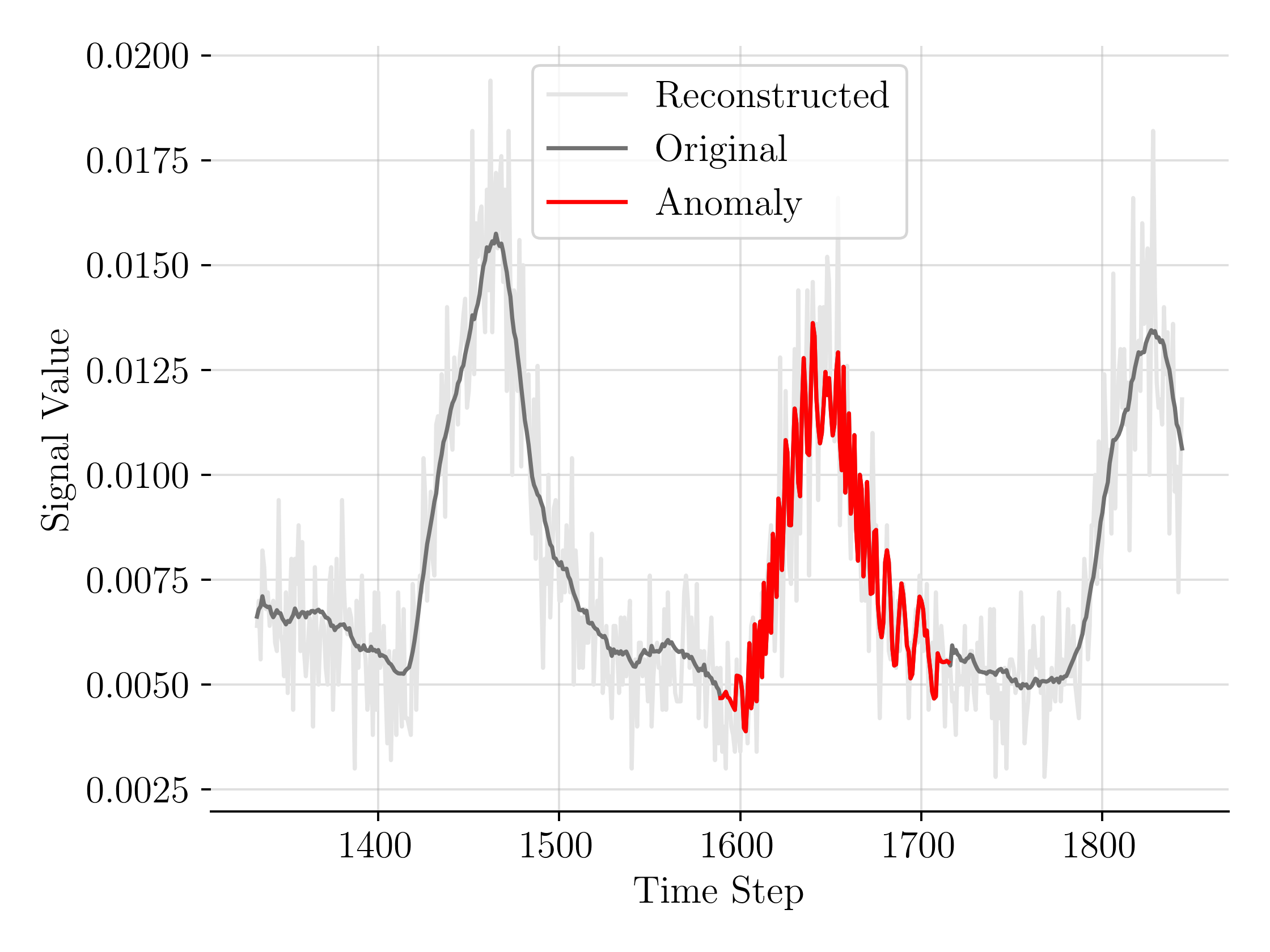}
    \caption{Illustration of a subset of the test data for dataset 28, with the reconstructed time series in the background, based on the best performing ansatz with respect to training loss. The anomalous range is highlighted in red. This figure underscores the necessity of adequately reconstructing the time series for this dataset, as the quantum autoencoder produces a non-smooth reconstruction of the original series that closely resembles the anomaly itself.}
    \label{fig:Set128AnomalyRecon}
\end{figure}

\textbf{The results in Table \ref{tab:AnomalyDetectionPerformance} demonstrate that for each dataset, quantum autoencoders outperform the classical autoencoder.} However, performance across datasets and ansaetze differ greatly. Performance disparities range from not detecting the anomaly a single time for the MSE-based approach on dataset number 28 to correctly detecting the anomaly at least once for every setup on dataset 176 with the Swap-based classification approach. Additionally, it appears that anomalies are especially hard to detect in datasets 28 and 99. 

This variability in performance underscores the necessity of "hyperparameter" optimization, similar to traditional machine learning, as even minor parameter adjustments can significantly impact performance. For instance, on data set 28, a RealAmplitudes ansatz with SCA entanglement and 28 parameters could not identify the anomaly once out of the 5 executions but adding one more layer and having 35 trainable parameters resulted in correctly identifying the anomaly two times. 

In addition, we want to emphasize that the \textbf{quantum autoencoders are less complex in terms of trainable parameters (ranging from 21 to 77) than the classical autoencoder baseline with 4,686 parameter. Moreover, the quantum autoencoders required fewer epochs to train - 45 epochs for quantum autoencoders vs. 250 epochs for the classical baseline.}

For dataset 28 it is evident that only RealAmplitudes with SCA entanglement managed to identify the anomaly correctly. Furthermore, the best performance was realized with 35 trainable parameters. This corresponds to the setup where RealAmplitudes with SCA entanglement usually performed best in terms of training loss. 

This is further supported by Figure \ref{fig:Set28Sfig1} where it can be seen that the anomaly detection performance requires low train loss and performance quickly deteriorates for larger train loss. Figure \ref{fig:Set28Sfig2} showcases the Swap-Test measurements on the test data for the best performing ansatz with respect to training loss  with a moving average window size of 128 and the valid detection range highlighted. In this particular instance the anomaly clearly differs from benign data as expected.

\begin{figure*}
\begin{subfigure}{.33\textwidth}
  \centering
  \includegraphics[width=1\linewidth]{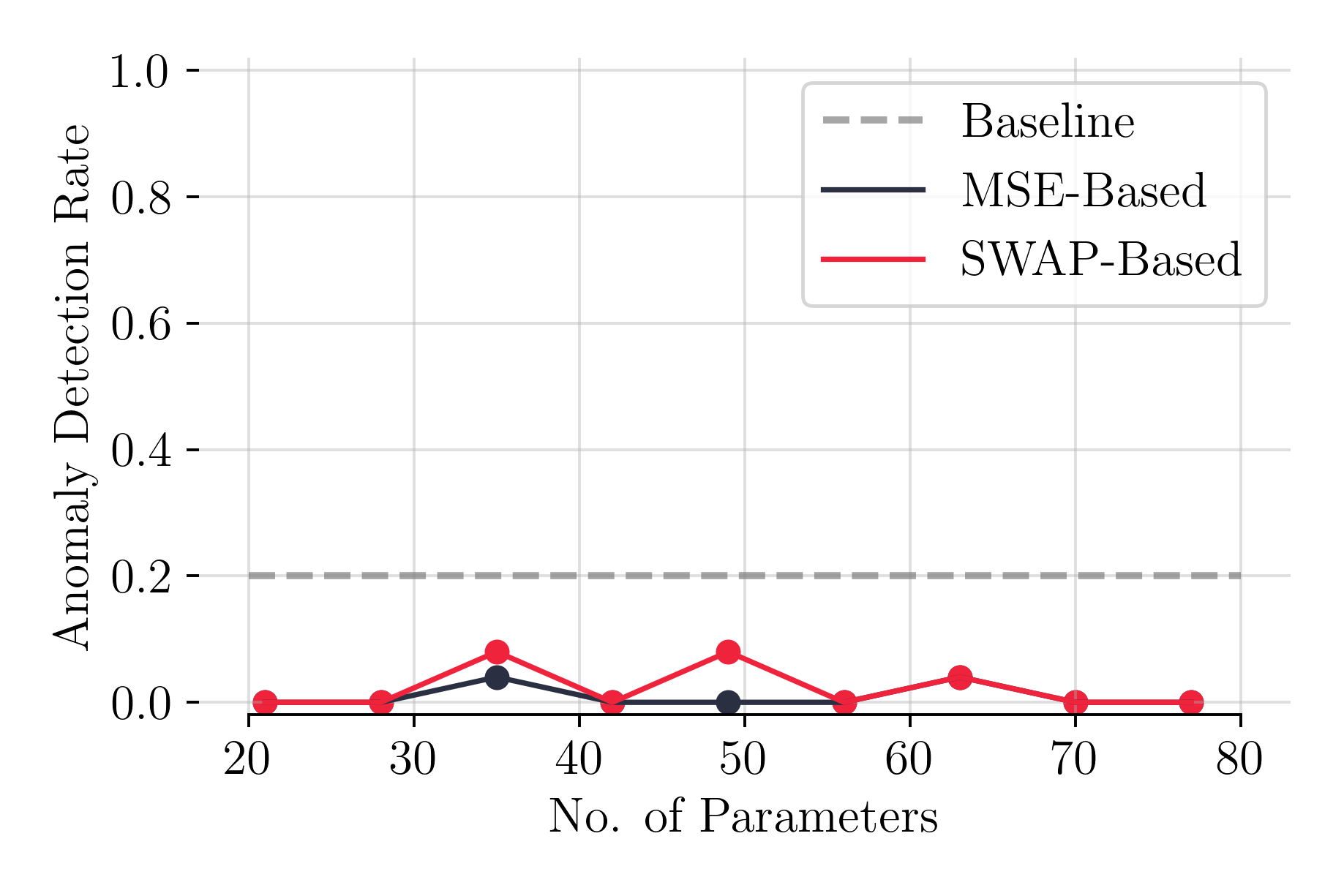}
  \caption{Dataset no. 28}
  \label{fig:MseVsSwapSfig1}
\end{subfigure}%
\begin{subfigure}{.33\textwidth}
  \centering
  \includegraphics[width=1\linewidth]{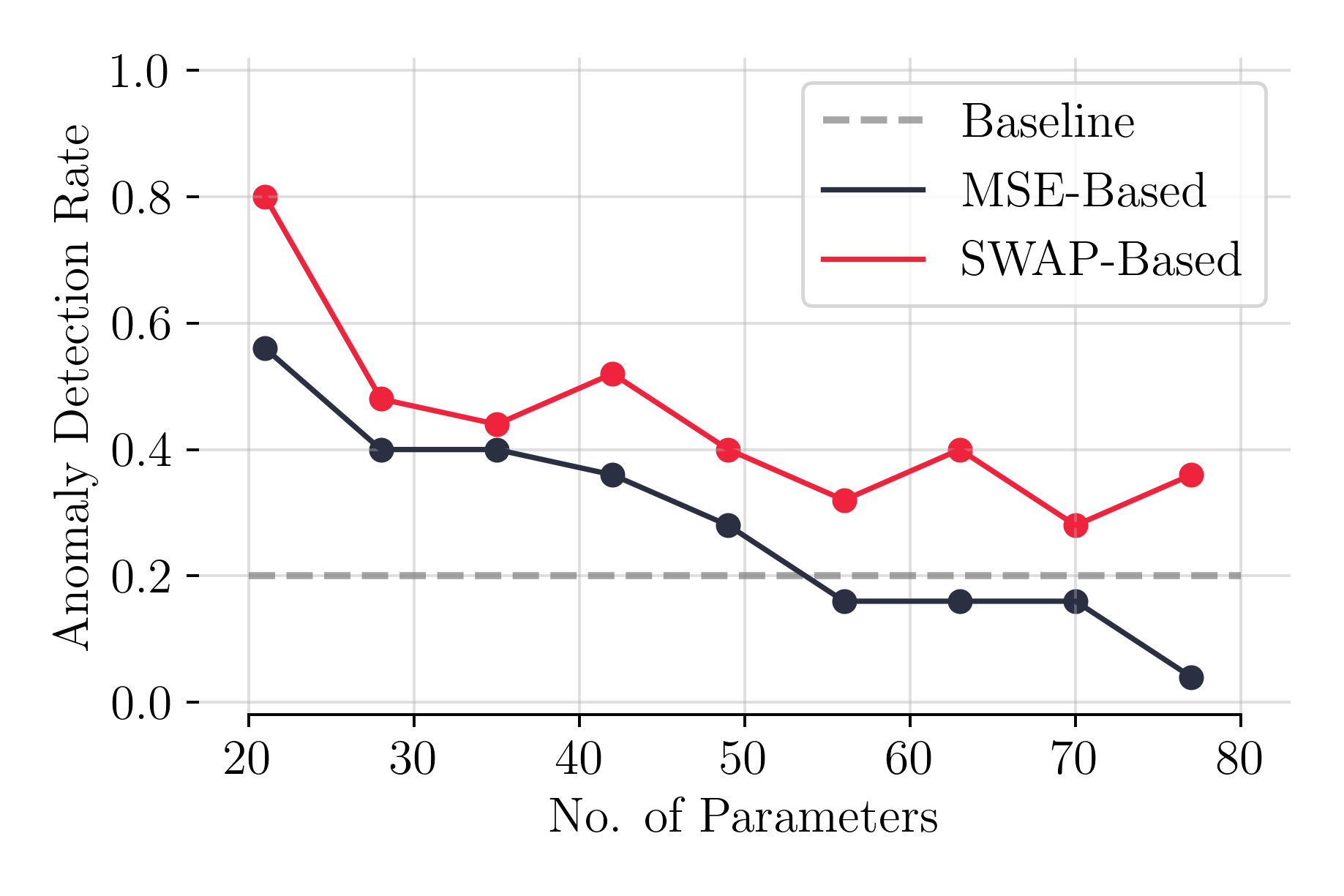}
  \caption{Dataset no. 54}
  \label{fig:MseVsSwapSfig2}
\end{subfigure}
\begin{subfigure}{.33\textwidth}
  \centering
  \includegraphics[width=1\linewidth]{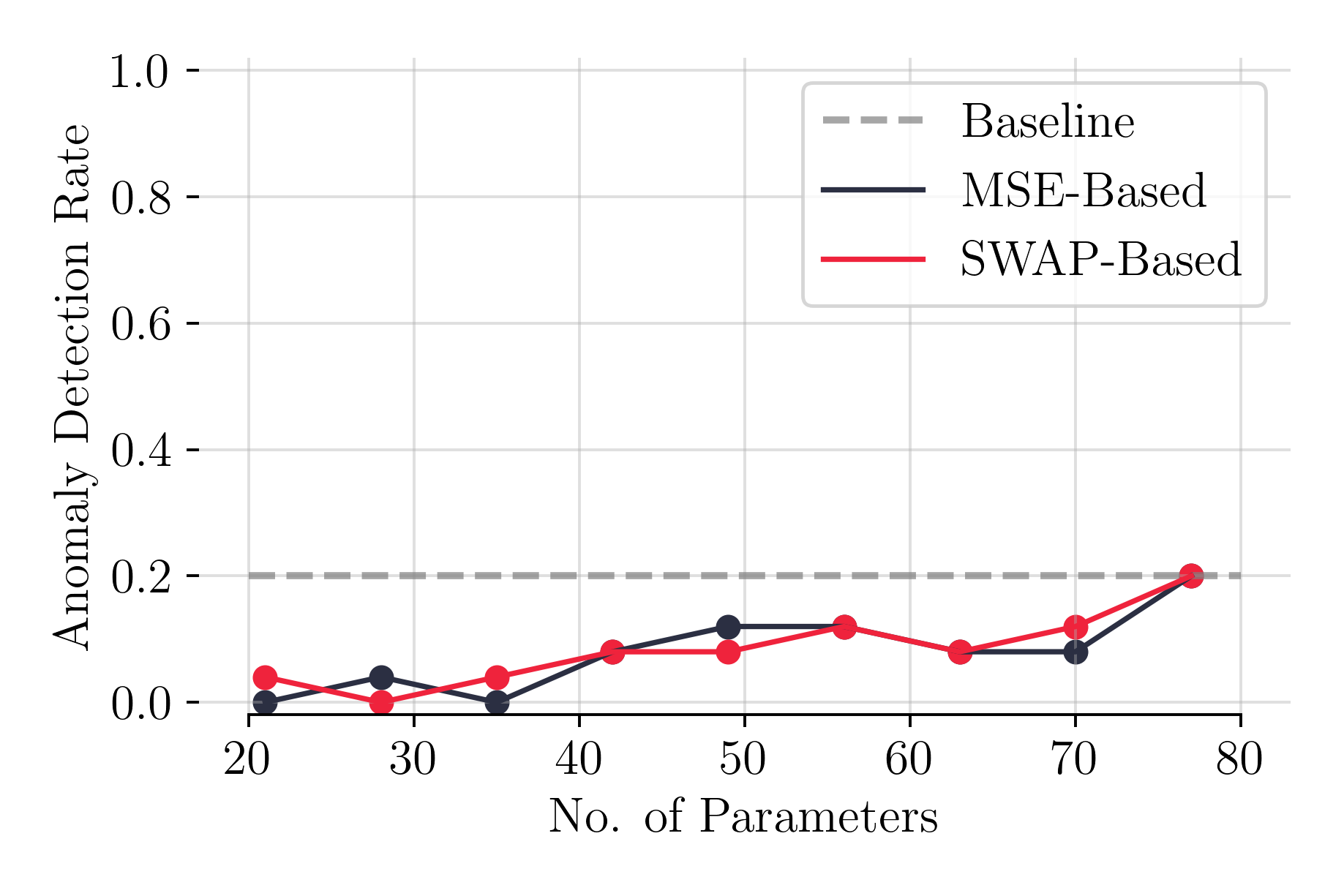}
   \caption{Dataset no. 99}
  \label{fig:MseVsSwapSfig3}
\end{subfigure}

\begin{subfigure}{.33\textwidth}
  \centering
  \includegraphics[width=1\linewidth]{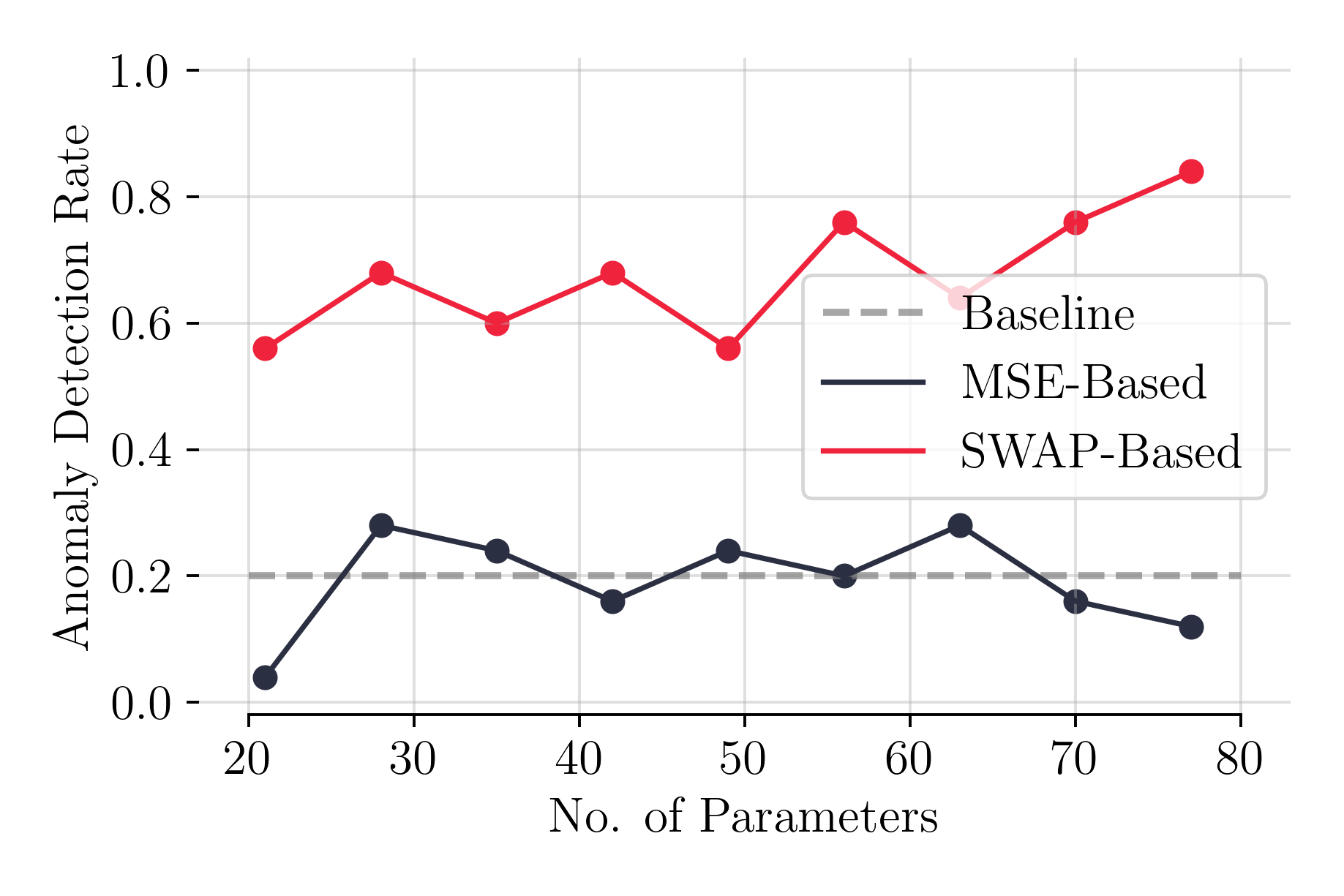}
  \caption{Dataset no. 118}
  \label{fig:MseVsSwapSfig4}
\end{subfigure}%
\begin{subfigure}{.33\textwidth}
  \centering
  \includegraphics[width=1\linewidth]{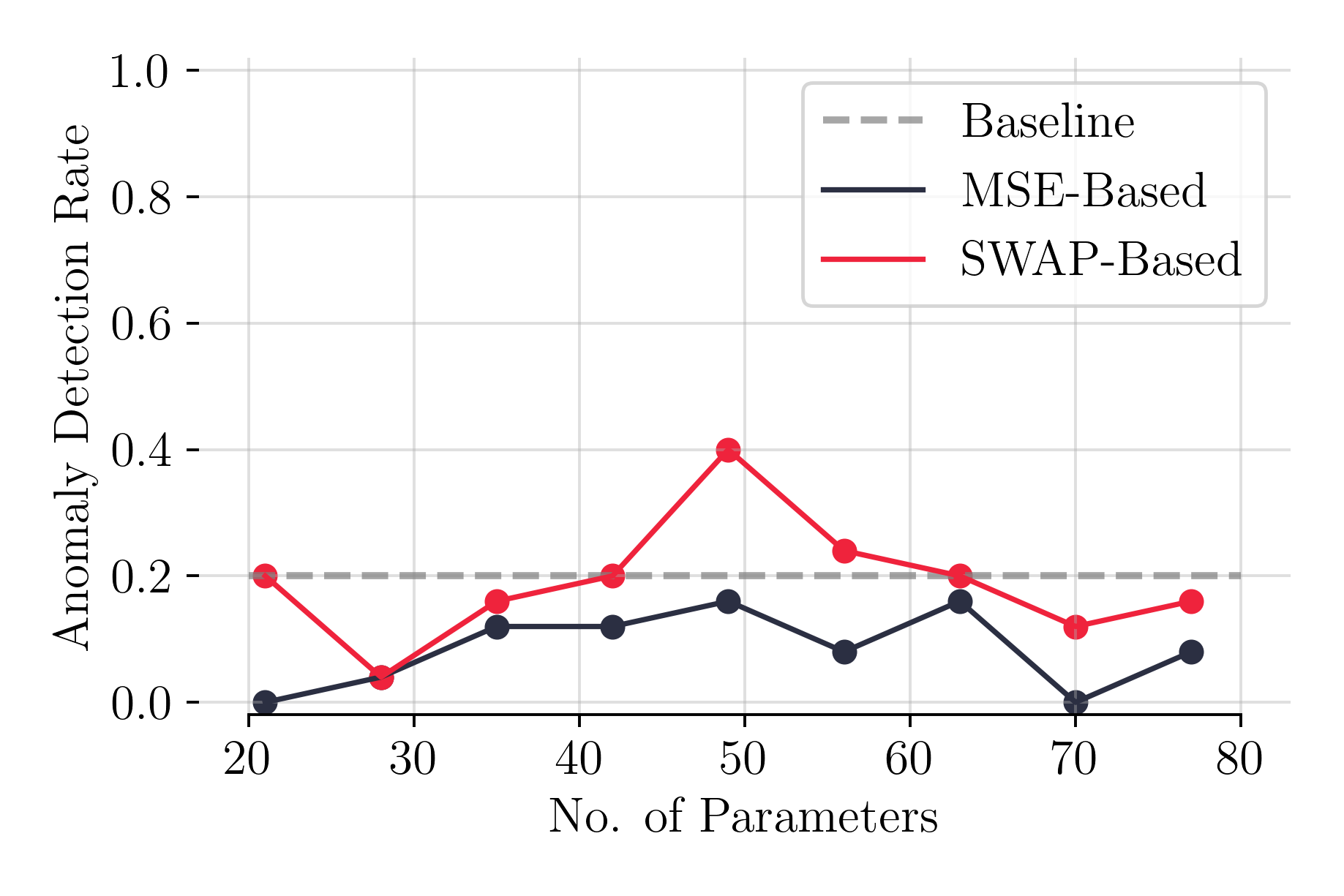}
  \caption{Dataset no. 138}
  \label{fig:MseVsSwapSfig5}
\end{subfigure}
\begin{subfigure}{.33\textwidth}
  \centering
  \includegraphics[width=1\linewidth]{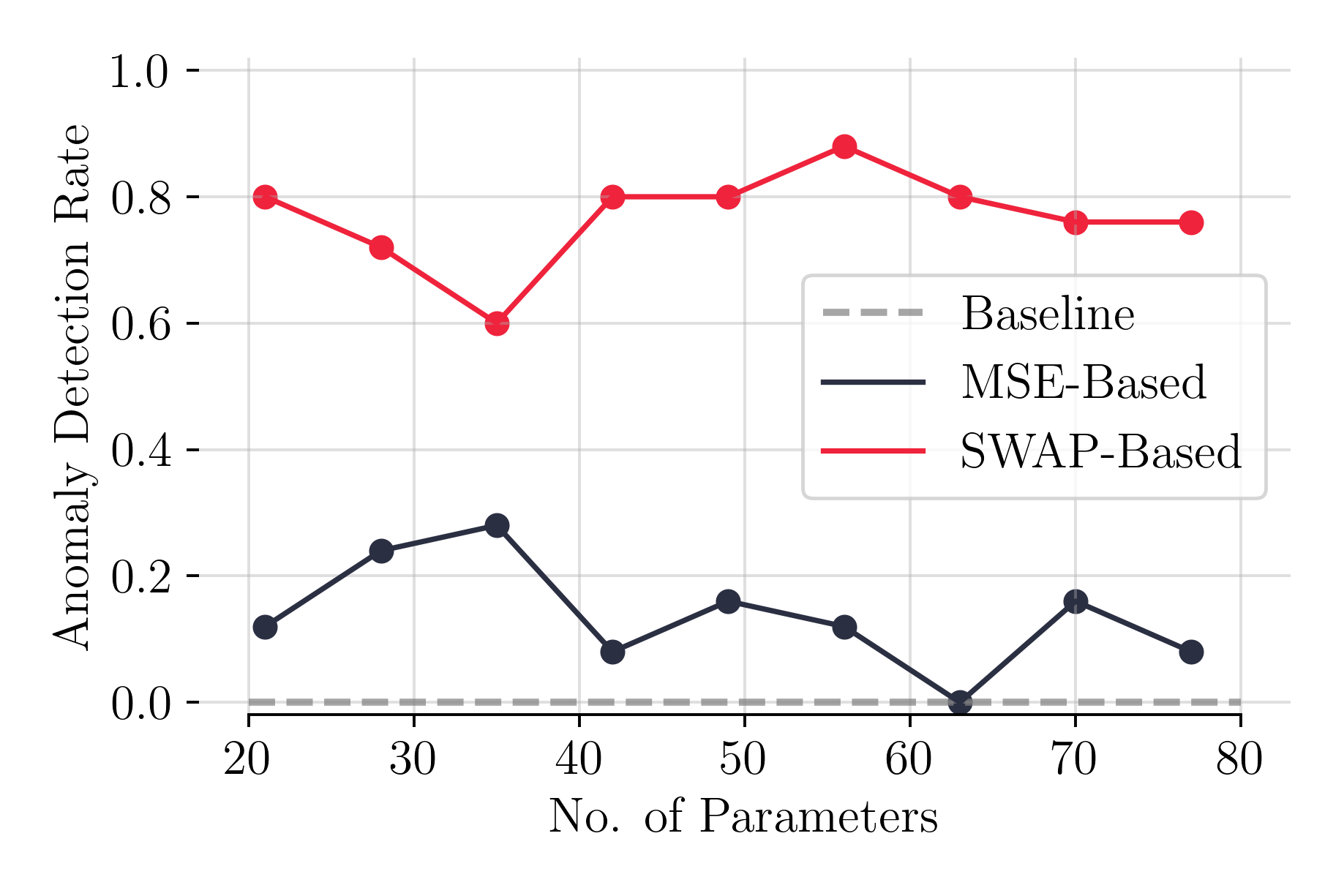}
  \caption{Dataset no. 176}
  \label{fig:MseVsSwapSfig6}
\end{subfigure}
\caption{This figure provides a comparative analysis of the anomaly detection capabilities for each detection approach. The comparison is based on detecting anomalies leveraging the Mean Squared Error (MSE) between the original and reconstructed time windows, and the classification of anomalies using the Swap-Test measurement. Values for both methods are averaged over all tested ansaetze at each number of parameters. A value of 1 indicates successful anomaly detection in every instance, while a value of 0 denotes complete failure to identify the anomaly. The baseline corresponds to the conventional deep learning-based autoencoder setup as described in Section \ref{sect:BaselineSetup}. Exhibiting subpar performance on a specific dataset compared to the baseline deep learning-based autoencoder does not imply that quantum autoencoders are intrinsically inferior for that dataset. Rather, it indicates that the average performance across all tested ansätze is lower. There may still exist a particular ansatz that surpasses the baseline, as demonstrated in Table \ref{tab:AnomalyDetectionPerformance}.}
\label{fig:MseVsSwap}
\end{figure*}

A possible explanation for this might be found in Figure \ref{fig:Set128AnomalyRecon} where a subset of the original as well as the reconstructed series can be found. It can be seen that the reconstruction is not as smooth as the original series. Paired with the noisy nature of the anomaly in this particular case, the reconstructed series resembles the anomalous part better than the benign data. It appears that in this case the quantum autoencoder needs to encode the training data extremely well in order to not produce such non-smooth reconstructions.

Notably, the difference in performance between the quantum autoencoder and the classical autoencoder is most significant on dataset 176 with the latter being unable to successfully detect the anomaly. For both the Swap-Test-based as well as the MSE-based approach there exist plenty of ansaetze consistently outperforming the baseline. However, it appears that reconstruction based approaches struggle more to identify the anomaly correctly as the Swap-based method significantly outperforms both the baseline and the MSE-based quantum autoencoder which both rely on reconstructing the input. This might be attributed to the subtlety of the anomaly in this particular dataset. 

As seen in Figure \ref{fig:DatasetOverviewSfig6}, the anomaly is hard to spot as it is very similar to the original series and only differs by being slightly noisy. Methods relying on reconstruction thus have to reproduce the input exceptionally well and require to be highly sensitive to small changes. The Swap-based methodology proves itself superior in this setting as it does not rely on adequate reconstruction of the input.

\subsubsection{Swap-based vs. MSE-based Detection}
\begin{figure*}
\centering
\begin{subfigure}{.5\textwidth}
  \centering
  \includegraphics[width=1\linewidth]{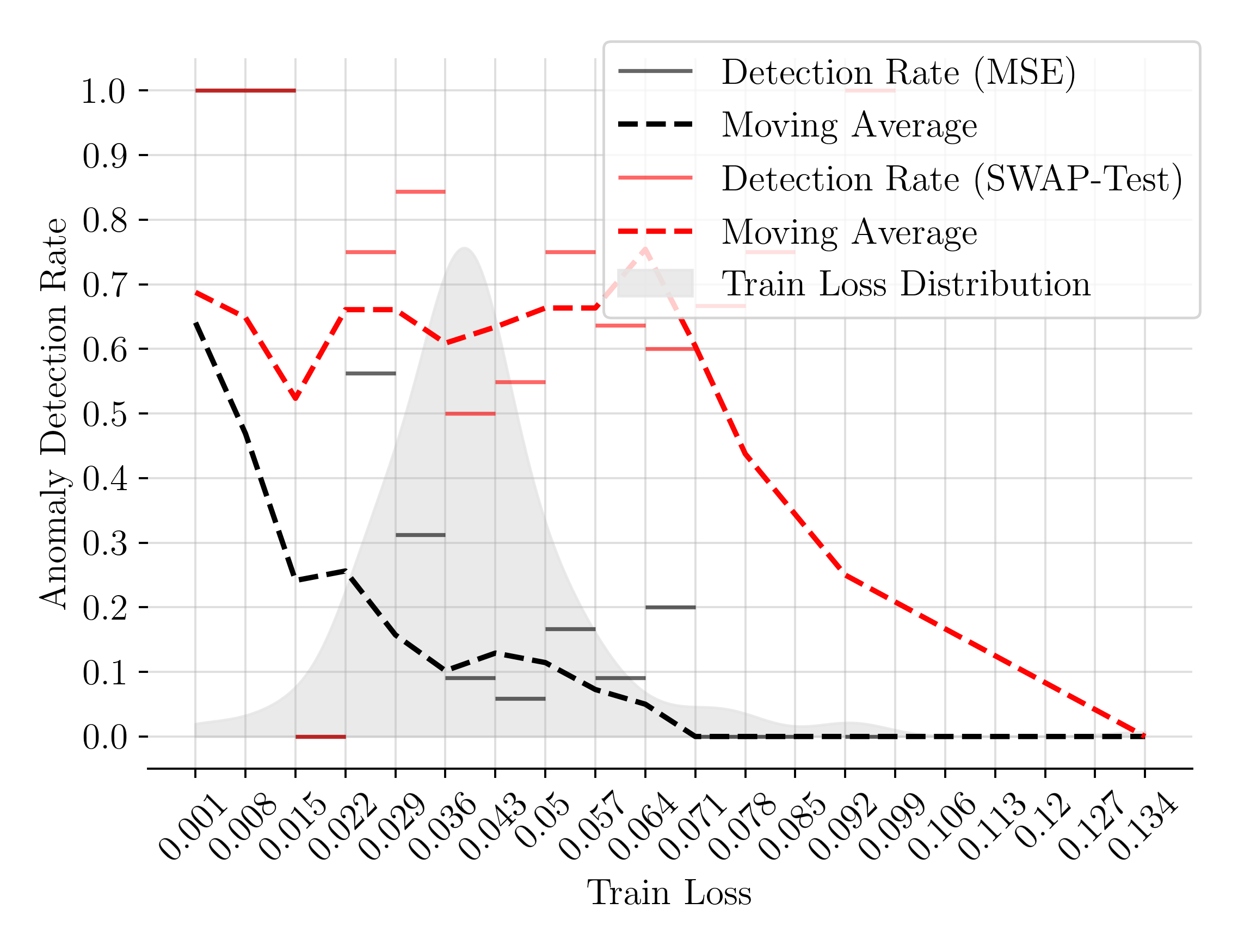}
  \caption{Dataset no. 118}
  \label{fig:MSEvSWAPvTrainLossSfig1}
\end{subfigure}%
\begin{subfigure}{.5\textwidth}
  \centering
  \includegraphics[width=1\linewidth]{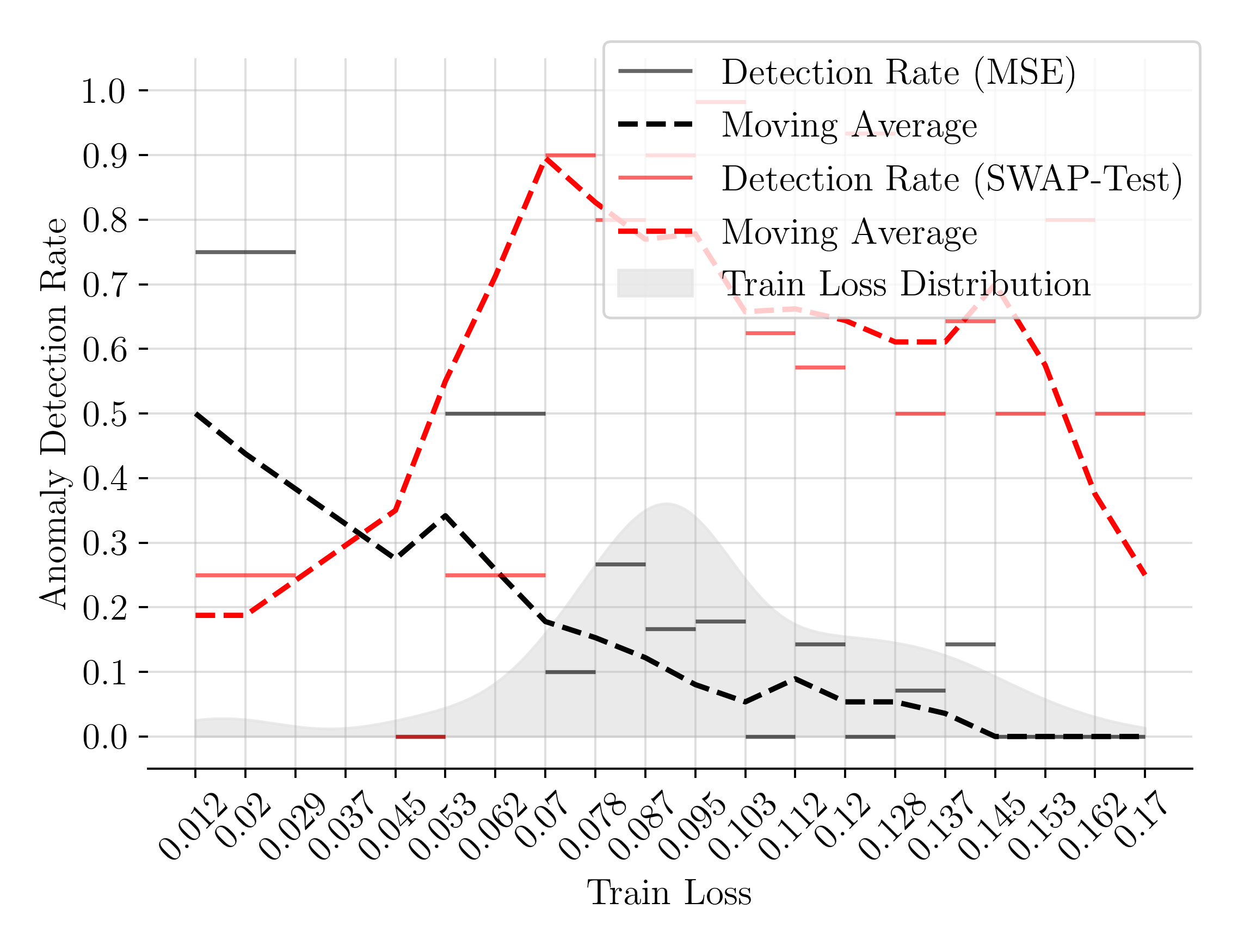}
  \caption{Dataset no. 176}
  \label{fig:MSEvSWAPvTrainLossSfig2}
\end{subfigure}
\caption{Subfigures \ref{fig:MSEvSWAPvTrainLossSfig1} and \ref{fig:MSEvSWAPvTrainLossSfig2} show the anomaly detection rate (the higher the better) for ansaetze with a given train loss falling in an interval of values for each the MSE based and Swap-Test based classification methodology. The detection rate is the average number of correctly identified anomalies for ansaetze showcasing a train loss falling into the associated interval. For instance, in Subfigure \ref{fig:MSEvSWAPvTrainLossSfig2} the MSE-based classification approach correctly identified 5 out of 6 anomalies with a train loss falling into the interval $[0.012,0.02]$. For the same quantum autoencoders, the Swap-Test based approach only identified the anomaly 2 out of 6 times. The moving average illustrates the trend of the anomaly detection capabilities of the two methodologies employed with increasing loss. The density in grey showcases how the train losses distribute for all ansaetze on the respective dataset highlighting where the majority of quantum autoencoders converge to.}
\label{fig:MSEvSWAPvTrainLoss}
\end{figure*}

Significant performance disparities between MSE-based and Swap-based anomaly detection are not limited to dataset 176, as shown in Figure \ref{fig:MseVsSwap}. For datasets like 118 and 176, where these discrepancies are pronounced, MSE-based classification often requires low training loss, as illustrated in Figure \ref{fig:MSEvSWAPvTrainLoss}. Specifically, for dataset 176 (Figure \ref{fig:MSEvSWAPvTrainLossSfig2}), MSE-based detection outperforms Swap-Test detection at low training loss, correctly identifying anomalies in 5 out of 6 cases within the $[0.012, 0.02]$ interval, compared to only 2 out of 6 for the Swap-Test.

In summary, Swap-Test-based anomaly detection appears more robust at higher loss levels, while MSE-based detection is more effective at lower loss. This pattern, along with the overall distribution of losses for the evaluated ansaetze (Figure \ref{fig:MSEvSWAPvTrainLoss}), may explain the superior performance of the Swap-Test approach on datasets 118 and 176, where most ansaetze converge to loss levels favoring Swap-based classification.



\subsubsection{Complexity Analysis}

\begin{figure*}
\centering
\begin{subfigure}{.5\textwidth}
  \centering
  \includegraphics[width=1\linewidth]{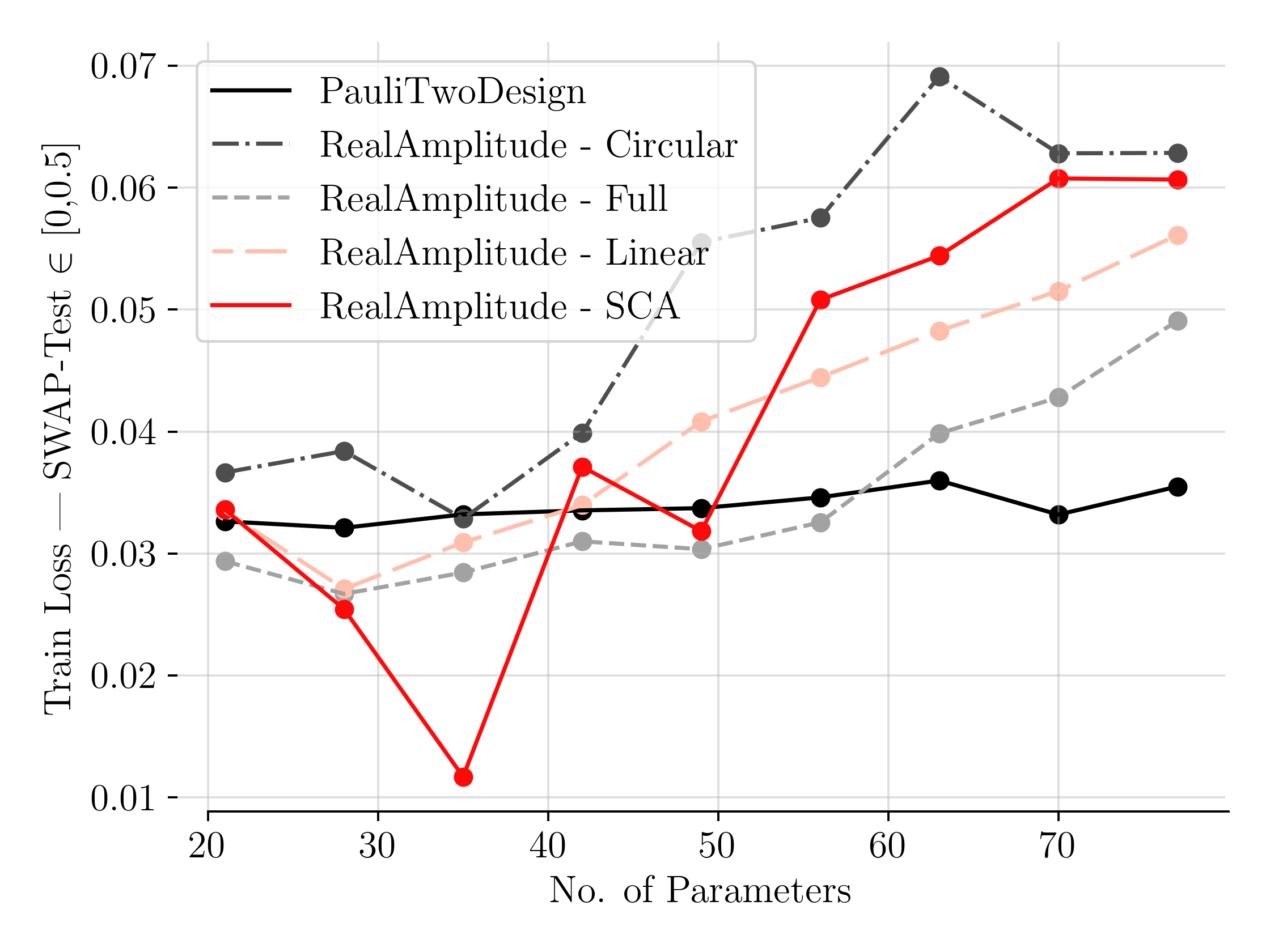}
  \caption{}
  \label{fig:ComplexityVsTrainingSfig1}
\end{subfigure}%
\begin{subfigure}{.5\textwidth}
  \centering
  \includegraphics[width=1\linewidth]{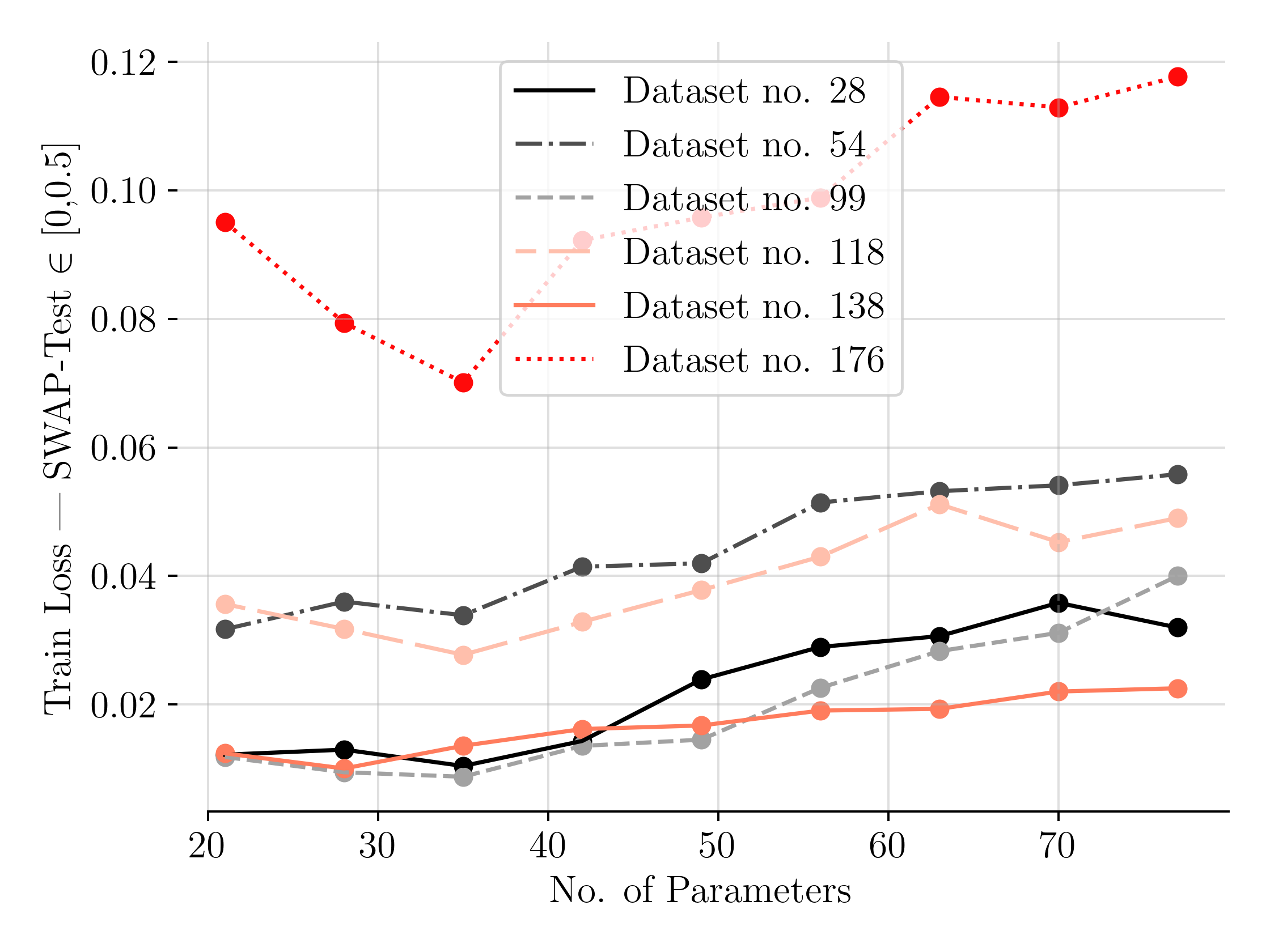}
  \caption{}
  \label{fig:ComplexityVsTrainingSfig2}
\end{subfigure}
\caption{Figures \ref{fig:ComplexityVsTrainingSfig1} and \ref{fig:ComplexityVsTrainingSfig2} depict the behavior of training performance as the number of trainable parameters increases. Figure \ref{fig:ComplexityVsTrainingSfig1} focuses on the ansaetze, showing the training loss at each parameter count as the average loss for that particular ansatz across all datasets. Conversely, Figure \ref{fig:ComplexityVsTrainingSfig2} presents the average training loss at each parameter count per dataset, with the averaged value derived from all ansaetze for a given dataset at a specific number of parameters. Generally, a low train loss is desired.}
\label{fig:ComplexityVsTraining}
\end{figure*}

We end by discussing the overall impact of the complexity of the chosen ansaetze on their training performance. Figure \ref{fig:ComplexityVsTraining} illustrates the behavior of training performance as the number of trainable parameters increases. Figure \ref{fig:ComplexityVsTrainingSfig1} focuses on the ansaetze, where the training loss at each number of trainable parameters represents the average loss for that particular ansatz across all datasets. Conversely, Figure \ref{fig:ComplexityVsTrainingSfig2} displays the average training loss at each number of parameters per dataset, with the averaged value derived from all ansaetze for a given dataset at a specific number of parameters.

It is evident that the training loss tends to increase with the number of parameters. This trend appears to be independent of the dataset. Notably, the RealAmplitudes ansatz combined with the SCA entanglement scheme demonstrates a significant drop in training loss at 35 parameters across all datasets. 

The phenomenon of increasing training loss for increasing complexity is counterintuitive, as one would typically expect better performance on training data with increased model complexity. However, this could be attributed to various factors, including the optimizer used in this study (COBYLA). Further research is necessary to determine whether this is a consequence of the training paradigm. 

The only exception to the increase in training loss poses the PauliTwoDesign where the train loss remains roughly constant with increasing complexity and might result from the fact that it was especially designed by \cite{McCleanBarrenPlateau} to overcome barren plateaus. 

To summarize, ansaetze relying on the RealAmplitudes design in  combination with COBYLA optimization tend to perform poorly with increasing number of parameters with the notable exception of RealAmplitudes in combination with the shifted circular alternation entanglement scheme and 35 parameters which significantly outperforms all ansaetze in terms of training performance. The PauliTwoDesign demonstrates near constant training performance across all complexities.

\begin{figure}[ht]
    \centering
    \includegraphics[width=1.0\linewidth]{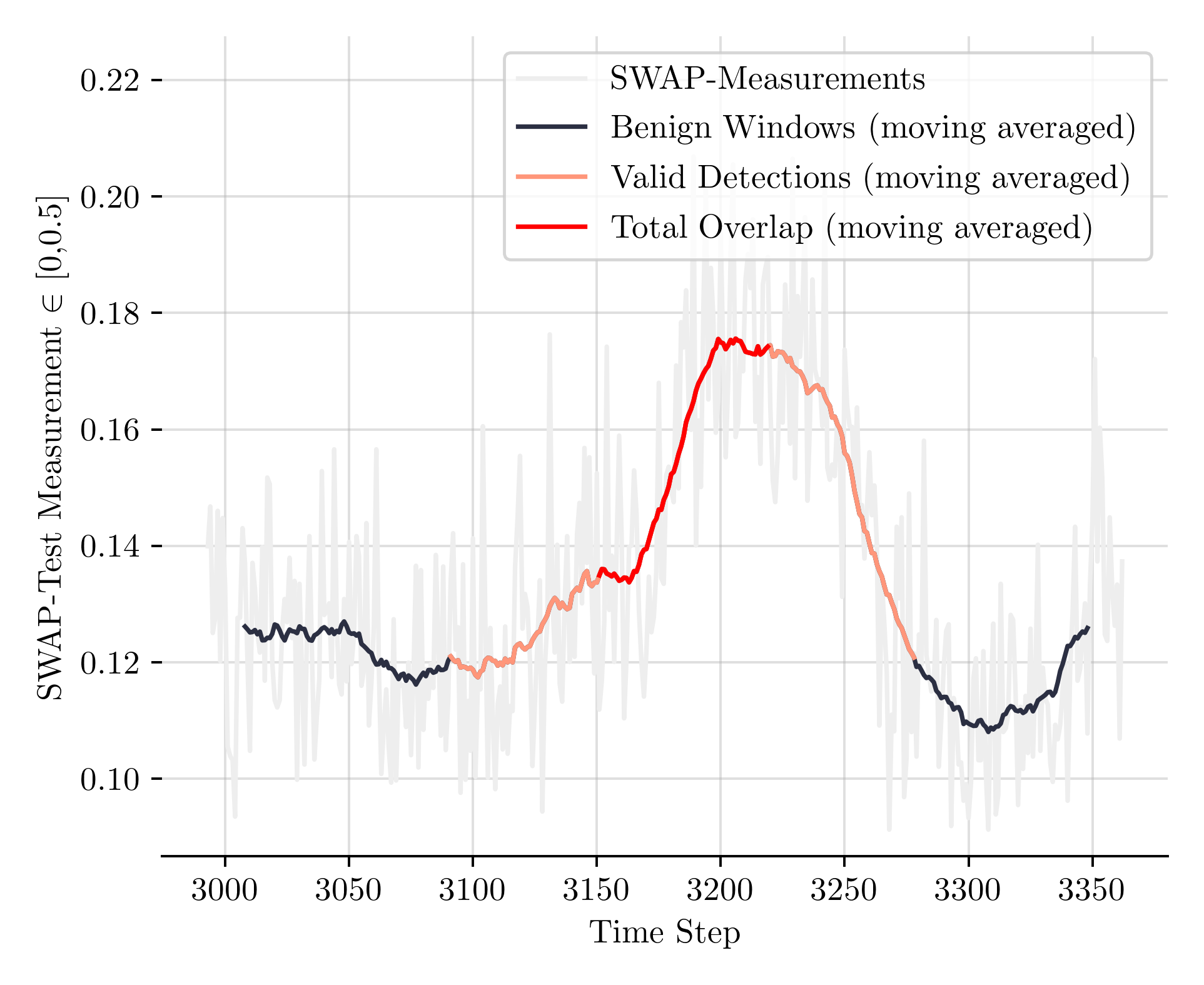}
    \caption{This figure illustrates Swap-Test measurements at $t$ for time windows from the test data starting at $t$ and spanning $128$ time steps, computed from the best performing quantum autoencoder trained on IBM Torino at the state of the best observed train loss. A moving average with a window size of 30 is applied. The "valid detection range", described in Section \ref{sect:UCRDataset}, indicates all time windows that contain anomalous time steps, while the "total overlap" identifies time windows fully encompassing the anomalous range. Larger Swap-Test measurements for the non-benign sections are desired in this instance.}
    \label{fig:SwapMeasurementAtT}
\end{figure}

\subsection{Results on Real Quantum Hardware}
\label{sect:RealHardwareResults}


The results in Table \ref{tab:QAERealHardwareOverview} show that \textbf{all quantum autoencoders, whether trained on a simulator or on IBM Torino, successfully identified the anomaly} by selecting a time window within the valid detection range.


Figure \ref{fig:SwapMeasurementAtT} presents the Swap-Measurement for the time window starting at $t_s$. Each value at $t_s$ corresponds to the Swap-Measurement obtained from the autoencoder, which encodes the time window starting at $t_s$ and spanning  $|W|$ time steps, where $|W|$ is 128 in this study. These measurements are derived from the best-performing quantum autoencoder trained on real hardware, specifically at the epoch exhibiting the lowest recorded loss, which in this case is epoch 25. The moving average used a window size of 30.

The similarity between measurement values at the start of the valid detection range (highlighted in orange) and those of non-anomalous values can be explained by the dataset characteristics discussed in Section \ref{sect:UCRDataset}. Early time windows in this range typically have minimal overlap with actual anomalies and consist mostly of benign values. This is supported by the increase in Swap-Measurements toward the middle of the range, where time windows fully capture the anomaly. The moving average smoothing process reveals that anomalous time windows, on average, produce larger Swap-Measurements, indicating the autoencoder's difficulty in accurately representing these anomalies within the 6-qubit latent space.

\begin{figure}[ht]
    \centering
    \includegraphics[width=1.0\linewidth]{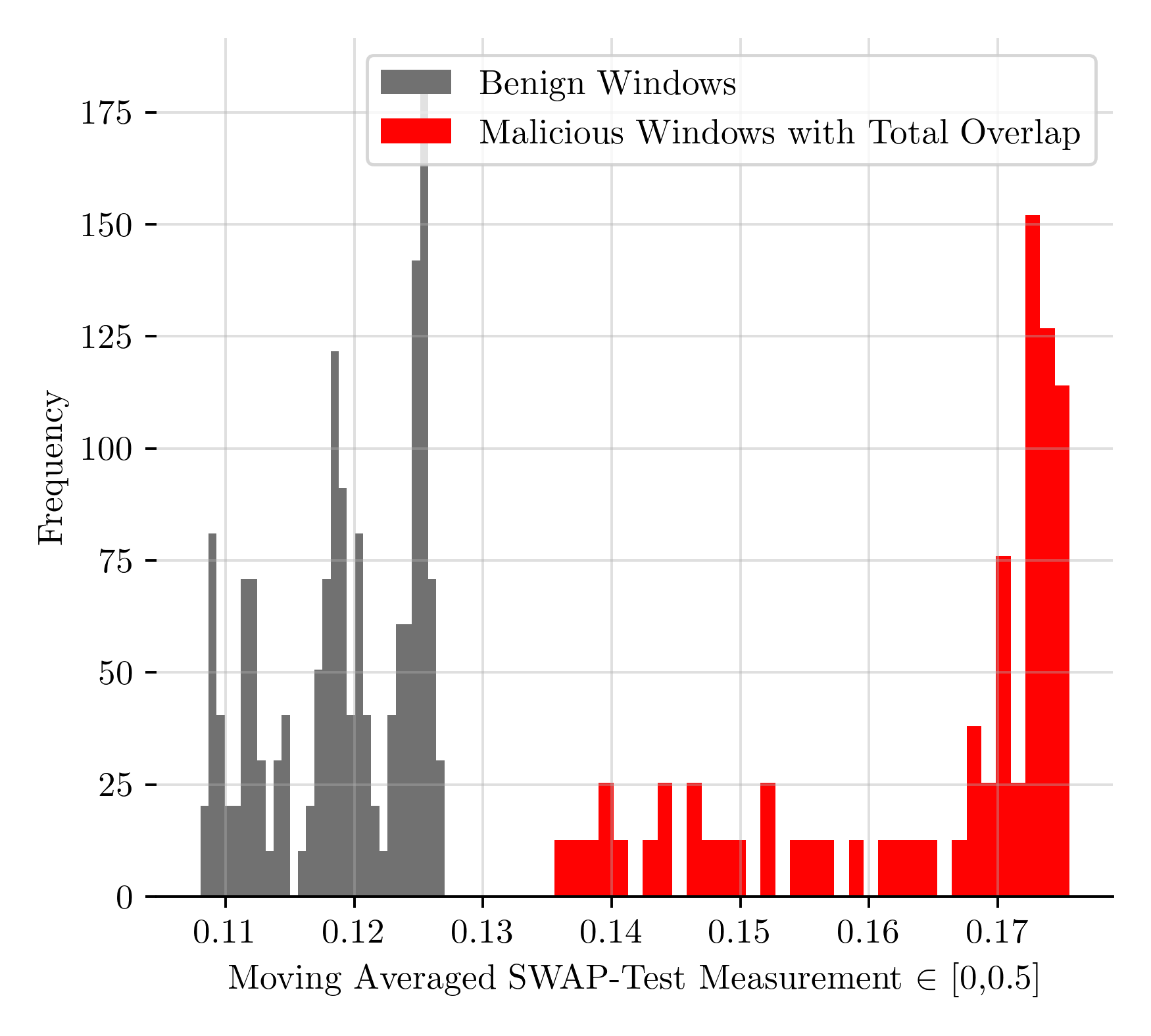}
    \caption{The figure presents distributions of post-processed Swap-Test measurements obtained from the top-performing quantum autoencoder on IBM Torino, corresponding to the state with the lowest observed training loss. The grey class represents benign time windows that do not overlap with the anomaly, while red comprises windows fully encompassing the anomalous range referring to values in the "Total Overlap" range in Figure \ref{fig:SwapMeasurementAtT}. Ideally, these distributions should exhibit minimal to no overlap, indicating that anomalies are well-separated from benign data. If the anomalous distribution is fully enclosed by the benign distribution, the anomaly becomes indistinguishable from normal data, rendering it unidentifiable.}
    \label{fig:SwapMeasurementsDistribution}
\end{figure}

Additionally, Figure \ref{fig:SwapMeasurementsDistribution} compares the distributions of post-processed Swap-Test measurements for both anomalous time windows fully encompassing the anomaly and benign time windows, further validating the quantum autoencoder's ability to distinguish between them. The benign distribution in Figure \ref{fig:SwapMeasurementsDistribution} corresponds to the moving averaged benign values (black solid line) in Figure \ref{fig:SwapMeasurementAtT}, while the malicious distribution aligns with the data marked in red.

\begin{table*}[ht]
\resizebox{1\linewidth}{!}{
\begin{tabular}{@{}lcccccccccccc@{}}
\toprule
No.                                                                                                                                & 1                  & 2                 & 3                 & 4                                      & 5                     & 6                    & 7                    & 8                                        & 9                       & 10                      & 12                     & 13                     \\ \midrule
\multicolumn{1}{l|}{}                                                                                                              & \multicolumn{4}{c|}{\begin{tabular}[c]{@{}c@{}}Quantum Autoencoder (QAE)\\ IBM Torino\end{tabular}} & \multicolumn{4}{c|}{\begin{tabular}[c]{@{}c@{}}Quantum Autoencoder (QAE)\\ Statevector Simulator\end{tabular}} & \multicolumn{4}{c}{\begin{tabular}[c]{@{}c@{}}Classical Autoencoder (AE)\\ (Baseline)\end{tabular}} \\
\multicolumn{1}{l|}{\begin{tabular}[c]{@{}l@{}}$ \varnothing$ Train Benign Metric\\ (QAE: SWAP-Measurement, AE: MSE)\end{tabular}} & 0.131              & 0.074             & 0.155             & \multicolumn{1}{c|}{0.131}             & 0.0577                & 0.07                 & 0.07                 & \multicolumn{1}{c|}{0.05}                & 0.000063                & 0.000029                & 0.000028               & 0.000055               \\
\multicolumn{1}{l|}{\begin{tabular}[c]{@{}l@{}}$ \varnothing$ Test Benign Metric\\ (QAE: SWAP-Measurement, AE: MSE)\end{tabular}}  & 0.118              & 0.115             & 0.123             & \multicolumn{1}{c|}{0.121}             & 0.047                 & 0.064                & 0.065                & \multicolumn{1}{c|}{0.042}               & 0.000064                & 0.000029                & 0.000029               & 0.000056               \\
\multicolumn{1}{l|}{Distribution Overlap}                                                                                          & 0.619              & 0.503             & 0.393             & \multicolumn{1}{c|}{0.212}             & 0.264                 & 0.329                & 1.0                  & \multicolumn{1}{c|}{0.283}               & 1.0                     & 1.0                     & 1.0                    & 0.987                  \\
\multicolumn{1}{l|}{\begin{tabular}[c]{@{}l@{}}Distribution Overlap\\ (Full Anomalous Range only)\end{tabular}}                    & 0.0                & 0.0               & 0.0               & \multicolumn{1}{c|}{0.0}               & 0.012                 & 0.0                  & 0.664                & \multicolumn{1}{c|}{0.0}                 & 0.748                   & 0.638                   & 0.729                  & 0.535                  \\
\multicolumn{1}{l|}{Anomaly Found}                                                                                                 & True               & True              & True              & \multicolumn{1}{c|}{True}              & True                  & True                 & True                 & \multicolumn{1}{c|}{True}                & False                   & False                   & False                  & False                  \\ \bottomrule
\end{tabular}
}
\caption{This table presents the experimental results of the quantum autoencoder executed on the quantum computer IBM Torino, the quantum simulator statevector simulator and a comparison against the classical autoencoders. All values are based on the parameters corresponding to the epoch with the lowest loss. The "Mean Train Benign  Metric" represents the observed train metric during the training phase and refers to the Swap-Test measurement for quantum autoencoders and to the mean squared error for the deep learning baseline. The "Test Benign Metric" indicates the performance of the respective approach at the point of optimal training loss, specifically evaluated on benign time windows within the test set and should be similar to the train metric. "Distribution Overlap" quantifies the degree of overlap between the distributions of the test metric (Swap-Test for quantum autoencoders and MSE for the baseline) for benign and malicious time windows. The "Full Anomaly Range Only Distribution Overlap" assesses this overlap exclusively considering all benign test data windows and those test windows that fully encompass the anomalous range. The less overlap the better benign and anomalous can be separated. Finally, the last row denotes successful identification of the anomaly. }
\label{tab:QAERealHardwareOverview}
\end{table*}

\paragraph{Separability of Anomalies}
To evaluate the efficacy of separating anomalous time windows from benign ones, we examine the data presented in Table \ref{tab:QAERealHardwareOverview}, which reports the fraction of overlapping post-processed measurements. This fraction is calculated by dividing the number of post-processed measurements of anomalous data that falls within the bounds of the distribution of benign measurements by the total number of anomalous measurements. This analysis is performed for all valid detection windows as well as for those fully overlapping with the anomalous region. This method corresponds to calculating the fraction of overlap depicted in Figure \ref{fig:SwapMeasurementsDistribution}, where, in this particular example, the distribution of Swap-Test measurements of benign time windows with those fully encompassing the anomalous region would result in zero overlap.

Despite quantum autoencoders trained on real quantum hardware exhibiting significantly larger loss values compared to their simulator-trained counterparts, as indicated by the first and second row in Table \ref{tab:QAERealHardwareOverview}, their ability to separate anomalous windows from benign ones shows relatively little difference. When comparing mean values and excluding the outlier for the simulated quantum autoencoder, the mean overlap for all anomalous windows is 0.43 for those trained on real quantum hardware and 0.29 for those trained on the simulator (Table \ref{tab:QAERealHardwareOverview}, third row). For windows fully overlapping with the anomalous region, the mean overlap is 0 for quantum autoencoders trained on IBM Torino and 0.004 for the simulated class (Table \ref{tab:QAERealHardwareOverview}, fourth row).

These findings suggest that, \textbf{quantum autoencoders trained on real quantum hardware, despite having almost twice the loss values of their simulated counterparts, perform slightly better in distinguishing between benign and anomalous data in this setup}. This observation raises the need for further verification to determine if this trend holds more broadly.

In comparison to the classical autoencoder, the superiority of the quantum autoencoders on this particular dataset is evident. Even when considering only the measurements \textbf{for time windows that entirely encompass the anomalous region, the classical autoencoder distributions display an overlap of approximately 70\%. In contrast, the quantum autoencoders trained on real quantum hardware exhibit zero overlap} (Table \ref{tab:QAERealHardwareOverview}, fourth row). Furthermore, when applying the defined classification mechanism, which involves flagging the time windows with the largest post-processed values (i.e Swap-Test measurements for quantum autoencoders and MSE for classical autoencoders) as anomalous, the classical autoencoders fail to correctly identify the anomaly.

\begin{figure}[ht]
    \centering
    \includegraphics[width=1.0\linewidth]{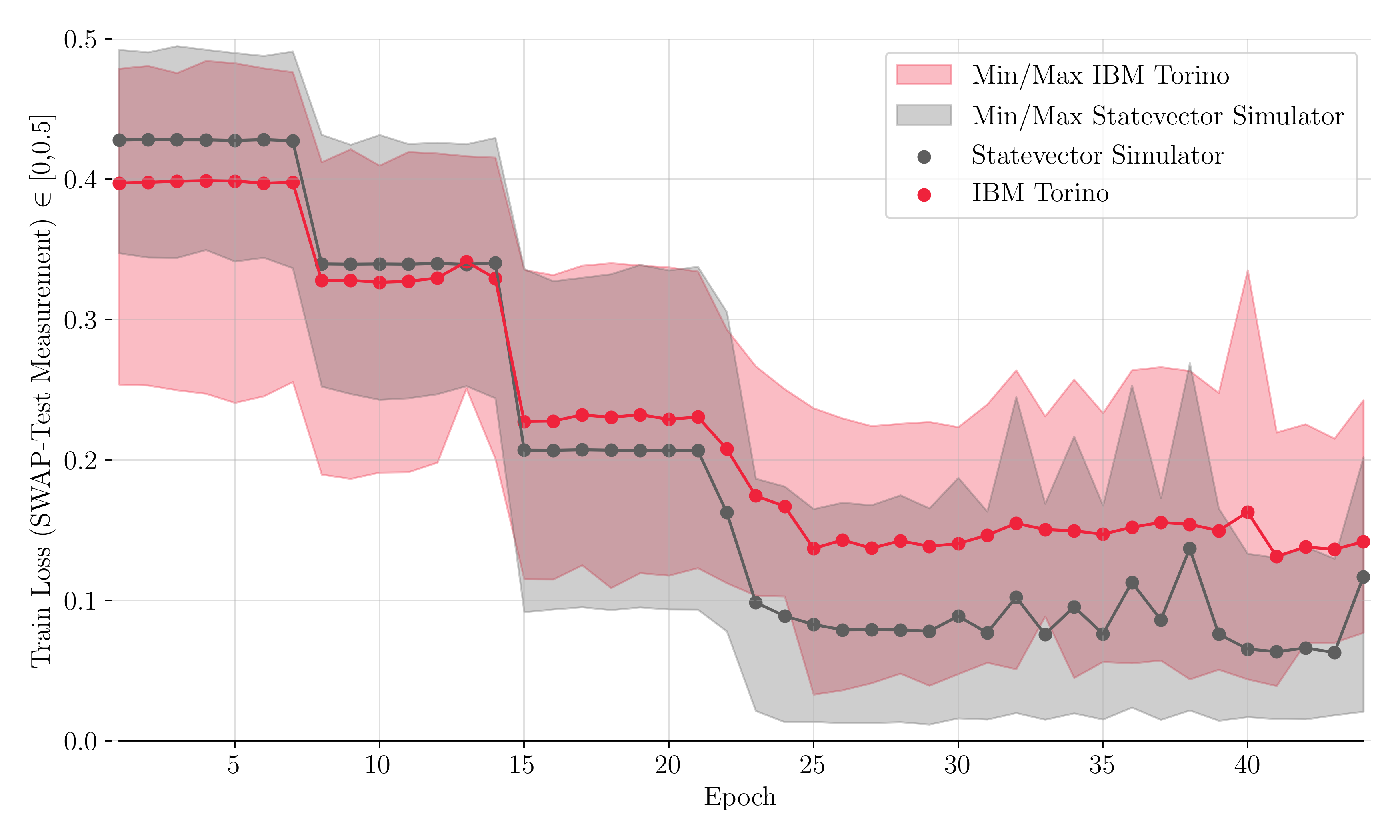}
    \caption{The mean loss curve derived from four executions of quantum autoencoders, comparing those trained on IBM Torino with those trained using the Statevector Simulator. The curves include the minimum and maximum loss values observed per epoch over all runs.}
    \label{fig:LossCurveQAE}
\end{figure}

\paragraph{Training Performance}
We conclude by examining the performance of quantum autoencoders trained on the IBM Torino quantum computer and the quantum simulator. Figure \ref{fig:LossCurveQAE} depicts the epoch loss for both setups, representing the average results of four executions, with the minimal and maximal values observed per epoch indicated.

The loss curves reveal that the quantum autoencoders trained on a simulator achieves a more rapid reduction in loss on average, whereas the quantum autoencoder exhibits a more gradual decline. By the conclusion of 45 epochs, both methods converge to similar loss values, although the simulator's final epoch shows a notable spike and generally maintains a lower loss throughout the training period. Additional epochs are required to effectivley determine whether both methods converge to identical loss values over an extended training period.

Each epoch required approximately 7 minutes of quantum computing time, totalling in a duration of about 11 minutes per epoch. Consequently, each experiment on real quantum computers consumed 5.25 hours of quantum computing time and took 8.25 hours to complete. This study required a total of 21 hours of quantum computing time, which would result in a cost of 120,960 USD, based on the pay-as-you-go pricing of 1.6 USD per second at the time of writing.

In this experimental setup, training the quantum autoencoder on the simulator demonstrates faster progress and superior training efficacy. The quantum autoencoder trained on real quantum hardware, however, shows enhanced robustness, with fewer erratic spikes in the loss curve. It is important to note that while the quantum autoencoder exhibits greater variability in loss values, this may be influenced by the limited number of repetitions (four) due to computational constraints.

\section{Related Work}
\label{sect:relatedWork}
The field of anomaly detection using quantum computing has gained considerable attention, but most research has focused on non-time series anomaly detection, employing a variety of approaches. Notably, \cite{HybridQMLAnomaly} and \cite{HybridQMLWang} explore hybrid methodologies, combining traditional deep learning with quantum computing. Specifically, \cite{HybridQMLAnomaly} propose a hybrid classical-quantum autoencoder, a modified deep learning autoencoder augmented with a quantum variational circuit in the latent space. Conversely, \cite{HybridQMLWang} replace individual layers in deep learning-based image classification, such as CNNs, with quantum variational circuits. Notably \cite{AnomDetectionHEP} employ a quantum autoencoder for anomaly detection in high energy physics and not only evalaute using simulators but also use real quantum hardware. However, the evaluated setup consisted of only two features.  

Additionally, \cite{QuantumCircuitAutoencoder} further demonstrate the anomaly detection capabilities of quantum autoencoders. However, the dataset utilized for this task does not adhere to commonly established benchmarks. Instead, it is generated by creating quantum states based on randomly chosen parameters for a given circuit. The benign and anomalous datasets are subsequently created by sampling parameters for the generating circuit from normal distributions $\mathcal{N}(0,0.1)$ and $\mathcal{N}(5,0.1)$, respectively. 

Further applications of quantum autoencoders were demonstrated by \cite{Zhu2023} where the authors employ quantum autoencoders to data compression in cloud-based processing. They successfully validate their approach on single- and multi-round communications and highlight the advantage of not revealing any server side computations other than the computed output. This specific use case further underscores the versatility and potential of quantum autoencoders.

For time series anomaly detection, \cite{bakerRewind} introduce Quantum Variational Rewinding (QVR). QVR trains parameterized unitary time-devolution operators to cluster normal time series data represented as quantum states. Anomaly scores are assigned based on the distance from the cluster center, and instances beyond a specified threshold are classified as anomalous. The approach is demonstrated on a simple case and applied to the real-world problem of detecting anomalous behavior in cryptocurrency market data. 

 \cite{QuantumTSClassification} apply quantum machine learning to time series classification through amplitude encoding. The methodology is rigorously evaluated using the well-established UCR time series classification benchmark. Nonetheless, their findings are derived exclusively from simulations, and the feasibility of their approach on actual quantum hardware remains unverified. Their work demonstrated that amplitude encoding of time windows constitutes an efficient method for representing a relatively large volume of data in relation to the number of qubits utilized and works well in the context of time series classification.

Regarding quantum autoencoders, \cite{RomeroQAE} laid the groundwork with their fully quantum autoencoder, which was subsequently extended by \cite{BravoQAE} and utilized in the present work. Although quantum autoencoders have shown promise in anomaly detection across various contexts, their application to time series anomaly detection using well-established and complex benchmarks, such as those from the UCR archive, remains unexplored. In this paper, we address this gap by applying quantum autoencoders to a challenging time series anomaly detection benchmark from the UCR archive. Previous studies have often utilized datasets that are either proprietary, lack complexity, or are not based on recognized benchmarks, limiting the broader applicability and validation of quantum autoencoders in real-world scenarios.

\section{Discussion and Conclusion}
\label{sect:DiscussionAndConclusion}
This study highlights the efficacy of quantum autoencoders in the domain of anomaly detection for time series data, demonstrating their good performance and significant potential to advance the field. \textbf{Our findings show that quantum autoencoders consistently outperform classical autoencoders across all tested datasets}. Notably, amplitude encoding emerges as a qubit-efficient method for data encoding in this context, facilitating effective anomaly detection with limited quantum resources. In addition, \textbf{quantum autoencoders required 60-230 times fewer parameters and 5 times fewer training iterations compared to the classical autoencoder}. However, the performance of quantum autoencoders is highly dependant on the chosen ansatz, emphasizing the critical importance of selecting an appropriate ansatz to optimize results.

\textbf{Furthermore, this study successfully demonstrates the implementation of quantum autoencoders on real NISQ quantum hardware}. Despite these promising findings, the application of \textbf{amplitude encoding on real NISQ hardware encounters significant limitations due to the extensive hardware requirements, which necessitate an impractically high number of quantum operations.} 

To address this challenge, the study investigates the use of quasi amplitude encoding as an alternative. Although this approach mitigates some hardware constraints, it remains feasible only for relatively small quantum systems due to its inherent limitations. Nonetheless, \textbf{even with quasi amplitude encoding, we successfully demonstrate the superior anomaly detection capabilities of quantum autoencoders compared to classical autoencoders, with the quantum autoencoder on real hardware matching the anomaly detection performance of the simulated one.} Furthermore, it is important to note that utilizing real quantum computer hardware with this approach demands a substantial amount of computational time, which may result in considerable costs. Therefore, cost considerations must be taken into account when employing such methods on NISQ quantum computers.

In conclusion, while quantum autoencoders in combination with amplitude encoding offer powerful tools for anomaly detection in time series data, further advancements in quantum hardware and encoding methods are essential for their broader applicability. \textbf{Future research should prioritize the development of more efficient ansatz designs and scalable encoding techniques to fully harness the potential of quantum computing in practical anomaly detection applications.}

Moreover, exploring anomaly detection within current limitations could involve extending the approach to multivariate time series. For instance, in this study, we analyzed univariate time series using 7 qubits, yielding time windows of 128 time steps. Increasing the qubit count would allow for processing 2-dimensional time series at equivalent resolution with a linear increase in complexity. 
Another potential improvement could be processing data at different resolutions simultaneously, as window size is often crucial for time series anomaly detection. 

\bmhead{Acknowledgements}
We would like to express our gratitude to Julien Baglio from QuantumBasel for his valuable contributions. We also extend our thanks to QuantumBasel for providing access to IBM Quantum Systems.

\bibliography{main}

\end{document}